\documentclass{article} 
\usepackage{iclr2026_conference,times}

\usepackage{amsmath,amsfonts,bm}

\def\eqref#1{equation~\ref{#1}}

\def\1{\bm{1}}

\DeclareMathAlphabet{\mathsfit}{\encodingdefault}{\sfdefault}{m}{sl}
\SetMathAlphabet{\mathsfit}{bold}{\encodingdefault}{\sfdefault}{bx}{n}

\usepackage{hyperref}
\usepackage{url}

\usepackage{booktabs}
\usepackage{multirow}

\usepackage[utf8]{inputenc}
\usepackage{booktabs}         
\usepackage[table]{xcolor}    
\usepackage{graphicx}         
\usepackage{amsmath}          
\usepackage{tcolorbox}        
\usepackage{array}            
\usepackage{enumitem}
\usepackage{tabularx}
\usepackage{float}
\usepackage{marvosym}
\usepackage{graphicx}
\usepackage{subcaption}
\newcommand{\best}[1]{\cellcolor{gray!20}\textbf{#1}}

\usepackage{fontawesome}
\usepackage{xspace}
\newcommand{\huggingface}{\raisebox{-1.5pt}{\includegraphics[height=1.05em]{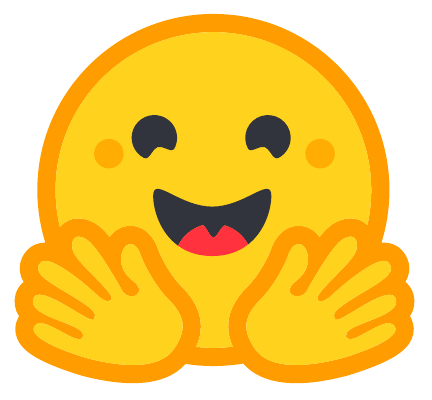}}\xspace}
\newcommand{\github}{\raisebox{-1.5pt}{\includegraphics[height=1.05em]{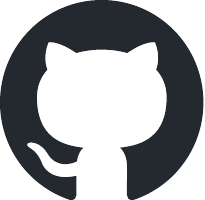}}\xspace}

\usepackage{xcolor} 

\definecolor{myblue}{HTML}{007FFF}
\definecolor{mygreen}{HTML}{009900}

\title{TIR-Bench: A Comprehensive Benchmark for Agentic Thinking-with-Images Reasoning}

\iclrfinalcopy

\author{
    Ming Li\textsuperscript{1}\thanks{Equal contribution.}\quad
    Jike Zhong\textsuperscript{2}\footnotemark[1]\quad 
    Shitian Zhao\textsuperscript{1}\footnotemark[1]\quad
    Haoquan Zhang\textsuperscript{1, 4}\footnotemark[1]\quad
    Shaoheng Lin\textsuperscript{1}\footnotemark[1]\quad
    Yuxiang Lai\textsuperscript{3}\footnotemark[1]\quad
 \\
 \textbf{Chen Wei}\textsuperscript{5}\quad
 \textbf{Konstantinos Psounis\textsuperscript{2}}\quad
    \textbf{Kaipeng Zhang\textsuperscript{1}}\thanks{Corresponding Author: zhangkaipeng@pjlab.org.cn} \\  
    \textsuperscript{1}Shanghai AI Laboratory\\
    \textsuperscript{2}University of Southern California\\
    \textsuperscript{3} Emory University\\
    \textsuperscript{4} Chinese University of Hong Kong\\
    \textsuperscript{5} Rice University\\
  \tt\small
  lm1640362161@gmail.com, zhangkaipeng@pjlab.org.cn 
}

\begin{document}

\maketitle
\vspace{-2em}
\begin{center} {\github \href{https://github.com/agents-x-project/TIR-Bench}{{\text{Implementation Code}}}}
\quad \quad
{\huggingface \href{https://huggingface.co/datasets/Agents-X/TIR-Bench}{{\text{Benchmark Data}}}}
\end{center}
\vspace{1em}
\begin{abstract}
The frontier of visual reasoning is shifting toward models like OpenAI o3, which can intelligently create and operate tools to transform images for problem-solving, also known as thinking-\textit{with}-images in chain-of-thought. Yet existing benchmarks fail to fully capture this advanced capability. Even Visual Search, the most common benchmark for current thinking-\textit{with}-images methods, tests only basic operations such as localization and cropping, offering little insight into more complex, dynamic, and tool-dependent reasoning.
We introduce \textbf{TIR-Bench}, a comprehensive benchmark for evaluating agentic thinking-with-images across 13 diverse tasks, each requiring novel tool use for image processing and manipulation in chain-of-thought. We evaluate 22 multimodal large language models (MLLMs), from leading open-sourced and proprietary models to those with explicit tool-use augmentation. Results show that TIR-Bench is universally challenging, and strong performance requires genuine thinking-with-images capabilities. Finally, we present a pilot study comparing direct versus agentic fine-tuning.
\end{abstract}

\section{Introduction}
The reasoning abilities of recent multimodal large language models (MLLMs)~\citep{OpenAI_GPT4o_2024,Gemini2p5} have advanced significantly, driven in large part by reasoning techniques such as chain-of-thought (CoT)~\citep{wei2022chain}. By decomposing reasoning of complex visual questions into a series of textual steps, MLLMs are able to achieve improved performance. 
While promising, these techniques are confined to the textual domain, conducting their reasoning solely through language while treating the visual information as a static, unalterable input~\citep{su2025thinking}.

To effectively process visual information, thinking-with-images has been proposed~\citep{thinkwithimage,su2025thinking,visualsketchpad}. This approach enables a model to generate new visual information by actively manipulating input images with tools. For example, when faced with a complex visual problem, OpenAI's o3 model~\citep{thinkwithimage} first writes code to create an image-processing tool, then executes it to modify the image (\textit{e.g.}, cropping, flipping, or rotating). The transformed visual data then informs the next stage of its linguistic reasoning.

To assess the agentic thinking-with-images capabilities of MLLMs, existing benchmarks~\citep{wu2024v,wang2025divide} have largely centered on visual search and tasks requiring the analysis of high-resolution images. These evaluations primarily validate a model's ability to accurately localize and crop specific regions within an image for better capturing detailed information to answer corresponding questions. However, these assessments tend to focus narrowly on the visual search capabilities of agentic MLLMs, leaving a broader spectrum of thinking-with-images abilities unevaluated. Therefore, there is an urgent need for a benchmark that incorporates a diverse range of tasks requiring sophisticated tool use to properly assess integrated multimodal reasoning.

In this paper, we introduce TIR-Bench, a comprehensive benchmark designed to evaluate the diverse thinking-with-images capabilities of agentic MLLMs. Unlike previous benchmarks focusing solely on visual search problems, TIR-Bench incorporates a diverse set of 13 tasks that require a wide range of tool-based interactions, such as zooming, rotating, increasing image contrast, adding auxiliary lines, and others to assess a model's tool integrated reasoning capabilities. The design of each task is predicated on the human intuition that solving it requires actively manipulating the visual input, rather than relying on static observation alone. For example, in TIR-Bench, a math problem might require the model to draw auxiliary lines or a coordinate system to find a solution, while a jigsaw puzzle task demands that it segment and then reassemble the image pieces. Consequently, TIR-Bench enables a more holistic evaluation of a model's thinking-with-images abilities, assessing a spectrum of skills not limited to visual search. 

\begin{figure}[t]
  \centering
  \includegraphics[width=\linewidth]{ 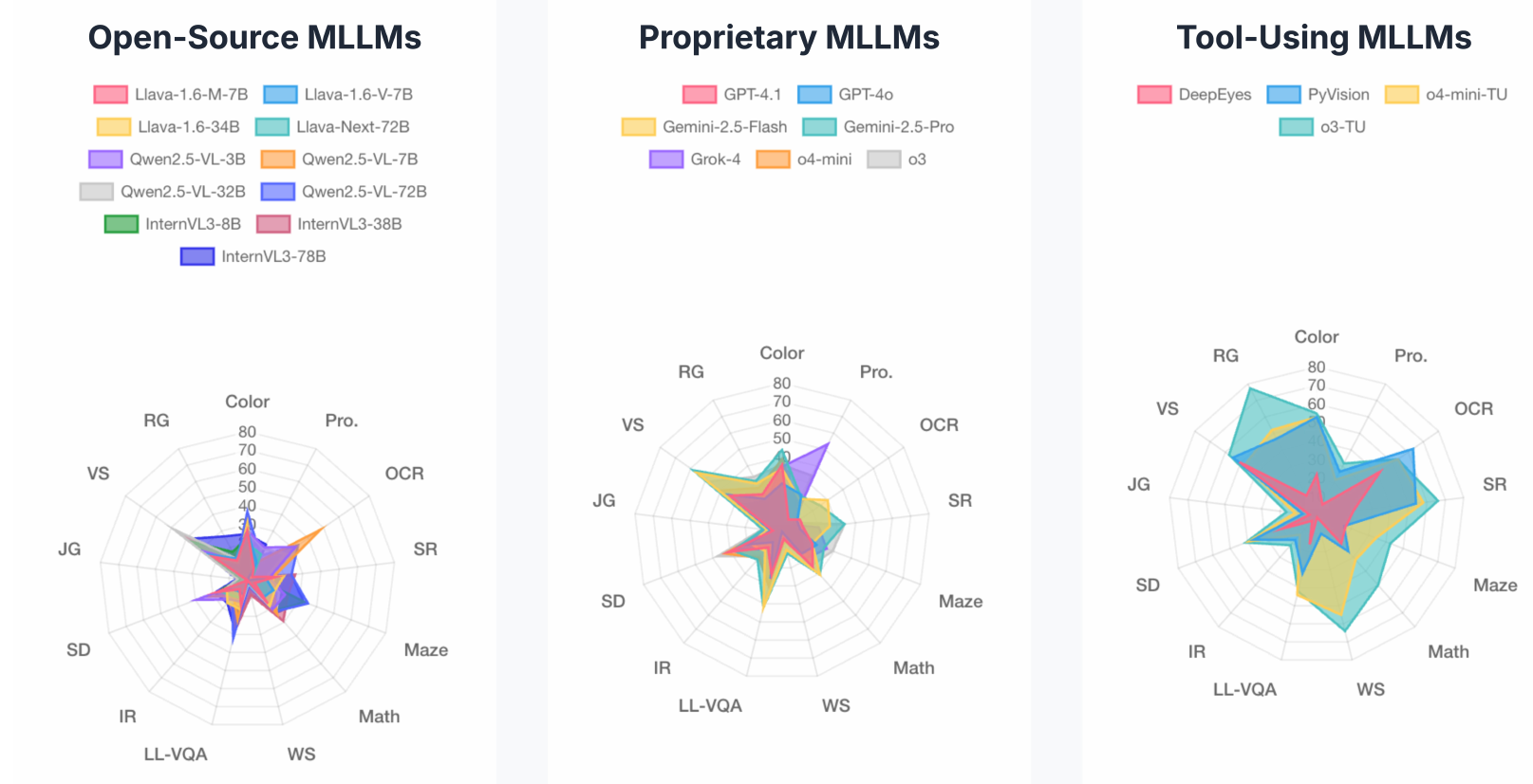}
  \caption{Overview of performance of open models (left), proprietary models (middle), and agentic tool-using models (right). SR: Symbolic Reasoning, WS: Word Search, LL-VQA: Low-Light VQA, IR: Instrument Reasoning, SD: Spot Difference, JG: Jigsaw Game, VS: Visual Search, RG: Rotation Game, Pro.: Proportion VQA. o3-TU: o3-tool-using, i.e., o3 with code interpreter. }
  \label{fig:f_radar}
  \vspace{-2em}
\end{figure}

Using TIR-Bench, we conduct a comprehensive performance evaluation of 22 leading MLLMs across three categories: open-source models, proprietary models, and tool-using agents. The overall experimental results, illustrated in Figure \ref{fig:f_radar}, reveal that TIR-Bench is a challenging benchmark for thinking-with-images abilities, as the best performance achieved is only 46\%. Moreover, traditional non-agentic models perform poorly on TIR-Bench, with the best-performing model Gemini-2.5-pro reaching an accuracy of merely 28.9\%. These findings highlight the importance of the thinking-with-images ability for this benchmark, as models equipped with tool-use capabilities, such as o3, o4-mini, and PyVision~\citep{zhao2025pyvision}, achieve much higher performance than other models.

Lastly, we assess the function-calling proficiency of various MLLMs and conduct a pilot study contrasting direct supervised fine-tuning~(SFT) with an agentic SFT approach on rotated image OCR task. Many recent MLLMs are equipped with function-calling capabilities. Our evaluation on rotation game task of TIR-Bench measures a model's proficiency in accurately executing the correct tool parameters as part of its reasoning chain. Results show that recent models like o3 perform well, whereas earlier models such as GPT-4o perform significantly worse. The pilot study on the rotated image OCR task compares two training methodologies and examines whether end-to-end SFT can achieve strong performance across different data scales for tasks involving image operations. Our findings indicate that the agentic SFT on full problem-solving trajectories with generated images is significantly more effective than direct SFT. This implies that agentic fine-tuning enables the emergence of more complex and robust problem-solving behaviors, allowing models to tackle multi-step tasks that are intractable with direct fine-tuning alone. 

\section{Related Works}
\subsection{Multimodal Benchmarks}
As the capabilities of Multimodal Large Language Models (MLLMs) evolve rapidly, a variety of benchmarks have been proposed to evaluate their performance, identify limitations, and guide future improvements~\citep{lu2023mathvista, zhang2024mathverse,qiao2024we,wang2024measuring,wang2024charxiv,li2024mmcode,lu2021inter,li2025eee}. These benchmarks are typically either specialized for specific domains~\citep{wang2024charxiv,li2024mmcode,lu2021inter,li2025eee}  or designed to be versatile and cover a broad range of tasks ~\citep{li2024seed,liu2023mmbench,yue2024mmmu,wang2023scibench}. MLLMs often use CoT reasoning on these benchmarks, though solving them only requires static image information. More recently, benchmarks such as V* Bench and HR-Bench have been introduced to evaluate agentic capabilities, specifically for visual search in high-resolution images~\citep{wu2024v,wang2025divide}. While these benchmarks advance agentic evaluation by requiring models to programmatically crop high-resolution images, their focus is narrowly confined to visual search. Consequently, a broader spectrum of thinking-with-images abilities, such as rotation and drawing, remains largely underexplored.

\subsection{Think with Images}

In previous works, Visual Sketchpad~\citep{visualsketchpad}, CoGCoM~\citep{cogcom}, DeepEyes~\citep{zheng2025deepeyes}, Pixel Reasoner~\citep{su2025pixelreasoner}, OpenThinkIMG~\citep{su2025thinking}, Chain-of-Focus~\citep{zhang2025chainoffocus}, Mini-o3~\citep{lai2025mini} and REVPT~\citep{zhou2025revpt} generate and executes tool calling in a predefined visual-specific toolset. In the intermediate reasoning steps, processed images are reinjected to the context, resulting in a multi-modal rational. However, these methods rely on a fixed collection of external visual parsers—such as detection models (e.g., GroundingDINO~\citep{groundingdino}) and segmentation models (e.g., SAM~\citep{kirillov2023sam})—which constrains their generality across diverse vision tasks and introduces bottlenecks due to dependency on external models. In contrast, ViperGPT~\citep{23viper}, o3~\citep{OpenAI_o3_2025}, o4-mini, Thyme~\citep{zhang2025thyme} and PyVision~\citep{zhao2025pyvision} adopt Python as its primitive tool. Capitalizing on the advanced coding and multimodal understanding capabilities of modern MLLMs—such as Claude-4.0~\citep{claude4} and GPT-4.1~\citep{gpt4.1}, enables the MLLM to dynamically write and execute code to construct complex, task-specific tools on demand, thereby supporting more general and flexible reasoning. This aligns with the emerging paradigm of “thinking with images” highlighted in o3’s blog~\citep{thinkwithimage} as a powerful cognitive capability. To assess this important ability, we propose TIR-Bench, covering diverse tasks on which multi-modal rationals are necessary.

\section{TIR-Bench}
\label{tir-bench}
\begin{figure}[t]
  \centering
  \includegraphics[width=\linewidth]{ 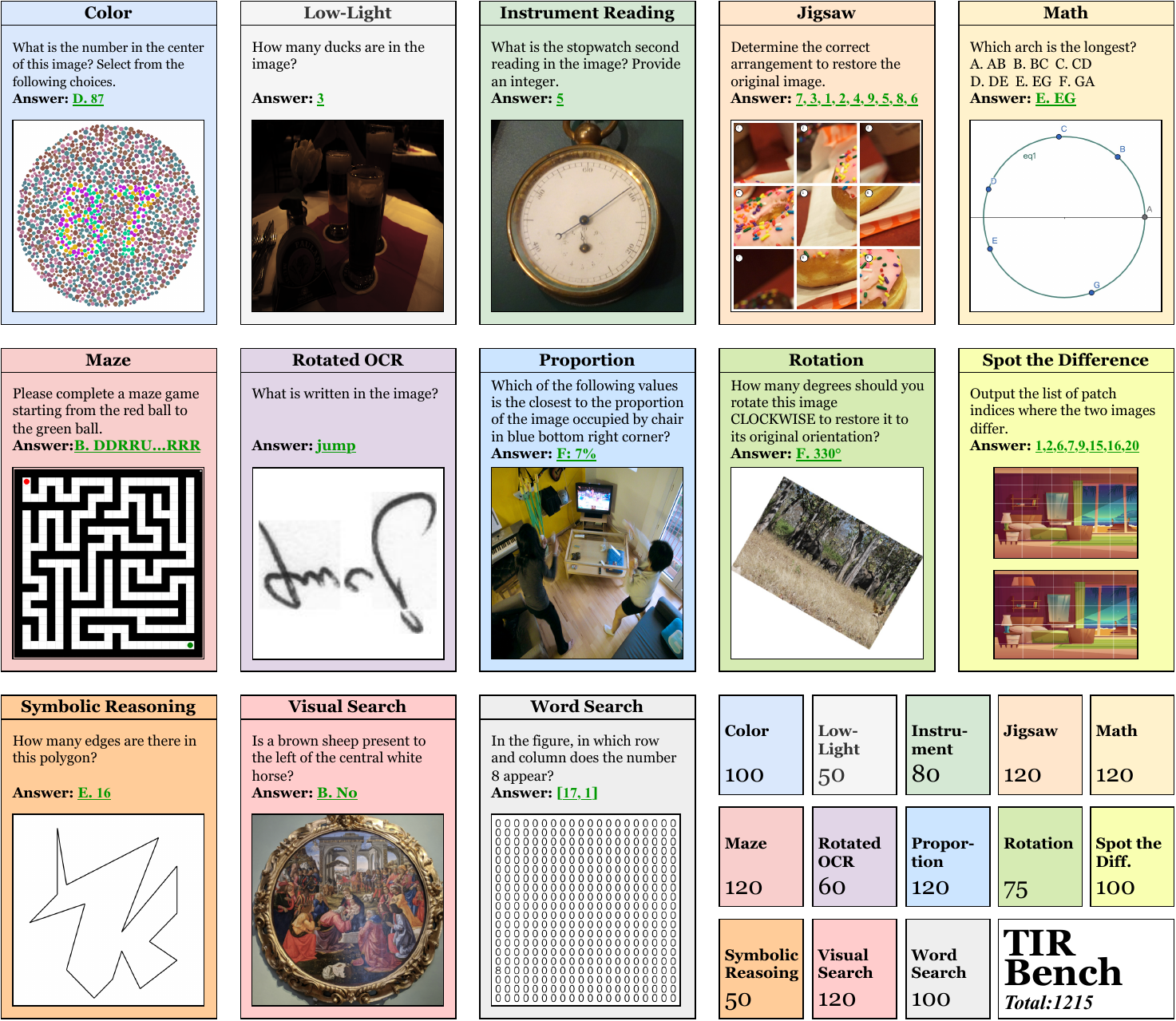}
  \caption{\textbf{Benchmark Overview.} TIR-Bench is composed of 13 tasks, meticulously designed to evaluate a wide spectrum of thinking-with-images capabilities.}
  \label{fig:demo}
\end{figure}

In this section, we introduce \textbf{TIR-Bench}. The overview of the benchmark is shown in Figure \ref{fig:demo}. We first introduce the task design strategy in \autoref{sec:task_design}. Next, we introduce the data collection process in \autoref{sec:data_collection}. Finally, we present the benchmark summary in \autoref{sec: summary}.

\subsection{Task Design}
\label{sec:task_design}
To extensively validate the model's ability to think with images we design 13 tasks. The benchmark's tasks are designed to evaluate a model's ability to perform active, tool-based visual reasoning, moving far beyond static image analysis. This includes tasks that require programmatic analysis, such as calculating color proportions or calling external models for object segmentation. Other challenges test the model's ability to overcome suboptimal conditions by programmatically enhancing low-light images or correcting the orientation of rotated text before performing OCR. The benchmark also features complex spatial and algorithmic puzzles, requiring models to solve mazes, reassemble jigsaw pieces, or draw auxiliary lines to solve geometric problems. Finally, it assesses fine-grained perception through tasks like spotting differences between images, reading instruments via cropping and zooming, and locating anomalies in visual puzzles. In every case, the model is forced to engage in a dynamic, multi-step process of visual manipulation and reasoning to arrive at the correct answer. More details about task design are introduced in \autoref{appendix:task}.
\subsection{Data Collection}
\label{sec:data_collection}
\textbf{Guidelines.} As previously mentioned, current benchmarks focus almost exclusively on visual search, overlooking a wide range of other tool-using abilities. To bridge this evaluation gap, we designed TIR-Bench in accordance with the following guidelines: (1) Diverse, Application-Grounded Tasks: TIR-Bench covers 13 distinct tasks across multiple domains, including spatial reasoning, visual perception, and mathematics, to mirror the complexity of real-world applications where static image analysis is insufficient. 
(2) Comprehensive Skill Assessment: The benchmark incorporates diverse visual contexts and requires a range of programmatic skills—from drawing auxiliary lines in geometry to executing pixel-level analysis—to foster a well-rounded evaluation of a model's agentic capabilities. 
(3) Probing Model Limitations: Each task is designed to be unsolvable without a multi-step, tool-based strategy. This intentional difficulty probes the limitations of current models, effectively distinguishing true thinking-with-images reasoning from simpler visual recognition. 
(4) Deterministic Evaluation: all tasks are designed with objectively verifiable answers, providing a robust framework for deterministic and reproducible evaluations.
(5) Many samples in TIR-Bench are newly annotated or generated, making it a more reliable benchmark that minimizes the risk of data contamination from models' pre-training corpora.

\textbf{Collection.} 
We briefly introduce the data collection process of the 13 tasks here, while the detailed process can be found in \autoref{appendix:task}. The summary of data collection is shown in Table~\ref{tab:task_comparison_grouped}.

\textbf{(1). Collection of Math VQA, Symbolic Reasoning, Low-Light VQA, and Instrument Reading tasks:} Data from these tasks are newly created and annotated data. We tasked two Ph.D. students with sourcing free images from the Internet and MathVista~\citep{lu2023mathvista} or create images using Geogebra, and then creating corresponding question-answer pairs. The annotators were explicitly instructed to design problems that, in their judgment, would necessitate tool-based image manipulation for a solution, ensuring the tasks could not be solved by static observation alone. 

\textbf{(2). Collection of Color VQA task:} We tasked a Ph.D. student with curating samples from the ColorBench~\citep{liang2025colorbench}, specifically selecting instances that could not be solved by static observation alone and therefore necessitate programmatic analysis to answer correctly. Finally, we select 100 problems from ColorBench~\citep{liang2025colorbench}.

\textbf{(3).  Collection of Jigsaw, Maze, Rotation, Word Search Tasks:} The data for these tasks were programmatically generated to create controlled and scalable challenges.
For the Jigsaw Puzzle, we selected 120 images from the RefCOCO dataset, chosen for their prominent objects. These images were then segmented into grids ranging from 3×3 to 6×6 and shuffled to create puzzles of varying difficulty. 
For the Maze task, we programmatically generated 100 mazes with sizes scaling from 5×5 to 62×62. 
For the Rotation Game, we utilized 75 images from CVBench~\citep{tong2024cambrian}, to which we applied random rotations. Three difficulty tiers were established based on the magnitude of the rotation angle (e.g., 5, 10, or 15 degrees). 
Finally, for the Word Search task, we programmatically generated 85 samples of different sizes and supplemented them with 15 complex puzzles sourced from the internet.

\textbf{(4). Collection of Spot the Difference:} We sourced pairs of nearly identical cartoon and real-world images from the internet, with one image in each pair containing subtle alterations. Both images were then segmented into an $m\times n$ grid of corresponding patches. Finally, two annotators reviewed the pairs to identify and label the specific patches that contained the differences.

\textbf{(5). Collection of Proportion VQA:} For this task, we collected 120 images from the RefCOCO dataset and used the ground-truth segmentation masks to calculate the correct object proportions. The incorrect multiple-choice options were then generated by adding or subtracting eight percentage points from the true value.

\textbf{(6). Collection of Rotated Image OCR Task:}
We selected 60 images from OCRBench and applied a rotation to each. The rotation probabilities were 25\% for 90°, 25\% for 270°, and 50\% for 180°.

\textbf{(7). Collection of Visual Search:} Recognizing that problems in existing benchmarks like V* Bench~\citep{wu2024v} are often too simple for current models, we curated a more challenging dataset for this task. We began by selecting 32 difficult problems from HR-Bench~\citep{wang2025divide}. To supplement these, we collected 88 new samples, which include 25 high-resolution art images and 63 high-resolution real-world images sourced from the internet. For each of these new images, an annotator was tasked with generating a unique question-answer pair. In total, the Visual Search task comprises 120 samples, 88 of which are newly created for this benchmark. 

\begin{table}[t]
\centering
\small

\resizebox{\textwidth}{!}{%
\begin{tabular}{l *{6}{l}}
\toprule
\textbf{Attribute} & \textbf{Color VQA} & \textbf{Proportion VQA} & \textbf{Symbolic Reasoning} & \textbf{Math} & \textbf{Word Search} & \textbf{Rotated OCR} \\
\midrule
Image Source & ColorBench & RefCOCO & VlmsAreBlind + Web-sourced & Web-sourced & Web-sourced & OCRBench \\
\midrule
QA Construction & Human-annotated & Synthetic & Human-annotated & Human-annotated & Synthetic & Dataset-provided \\
\midrule
Samples & 100 & 120 & 50 & 120 & 100 & 60 \\
\midrule
Answer Type & Single-choice & Single-choice & Single-choice & Single-choice & Open-ended & Open-ended \\
\bottomrule
\end{tabular}%
}

\vspace{0.4cm}

\resizebox{\textwidth}{!}{%
\begin{tabular}{l *{7}{l}}
\toprule
\textbf{Attribute} & \textbf{Maze} & \textbf{Low-Light} & \textbf{Instrument Reading} & \textbf{Spot the Difference} & \textbf{Jigsaw} & \textbf{Visual Search} & \textbf{Rotation} \\
\midrule
Image Source & Synthetic & Web-sourced & Web-sourced & Web-sourced & Synthetic & Web-sourced & CVBench \\
\midrule
QA Construction & Synthetic & Human-annotated & Human-annotated & Human-annotated & Synthetic & Synthetic & Synthetic \\
\midrule
Samples & 120 & 50 & 80 & 100 & 120 & 120 & 75 \\
\midrule
Answer Type & Single-choice & Open-ended & Open-ended & Open-ended & Open-ended & Single-choice & Single-choice \\
\bottomrule
\end{tabular}%
}

\caption{Overview of all tasks, grouped into reasoning-oriented (top) and perception-oriented (bottom), with their sources, QA constructions, sample sizes, and answer types.}
\label{tab:task_comparison_grouped}
\end{table}

\subsection{Benchmark Summary}
\label{sec: summary}
TIR-Bench consists of a total of 1215 examples divided
into 13 diverse but essential tasks. Questions in our benchmark are categorized into two types: multi-choice and free-form problems, counting 665 and 550, respectively. The average length
for question text is 46.29 and the average length for answer text is 3.41. The distribution of each task is shown in Figure~\ref{fig:task_dis}. Image and question examples for each task are shown in \autoref{appendix:task_example}.
\section{Experiments}
In this section, we detail our evaluation setup and results for \textbf{22} leading MLLMs, considering both models with and without agentic tool-use capabilities. Our goal is to show that MLLMs lacking the ability to think with images perform poorly on TIR-Bench. We describe the experimental setup in \autoref{exp-setup}, report the results in \autoref{exp_results}, present experiments on function calling in \autoref{exp_funciton}, and compare agentic SFT with end-to-end SFT in \autoref{exp_ft}.
\subsection{Experiment Setup}
\label{exp-setup}
\paragraph{Model selection.} We categorize the MLLMs we use into open-source, proprietary, and tool-using. For open-sourced, we evaluated 11 models across three widely used and up-to-date model families: LLaVA \citep{li2024llava, liu2024improved}, Qwen2.5-VL~\citep{bai2025qwen2}, and InternVL3~\citep{zhu2025internvl3}, ranging from 3B to 78B. Results from these models accurately reflect open-source MLLMs' performance on thinking-with-image reasoning tasks. For proprietary models, we selected 7 models across 3 model families: GPT \citep{OpenAI_GPT4o_2024}, Gemini-2.5 \citep{Gemini2p5}, and Grok-4 \citep{Grok4_xAI_2025}. We test GPT series, including GPT-4.1, GPT-4o \citep{OpenAI_GPT4o_2024}, as well as o-series models \citep{OpenAI_o3_2025}, including o3 and o4-mini. For these GPT models, we use Azure API for calling models w/o python interpreter or sandbox~\citep{thinkwithimage}. For agentic tool-using (TU) MLLMs, we evaluate three open-sourced frameworks: DeepEyes \citep{zheng2025deepeyes}, and PyVision \citep{zhao2025pyvision}, and 2 proprietary models o4-mini-TU and o3-TU. We use the official OpenAI API and turn on code interpreter and set the container parameter as auto.

\paragraph{Evaluation.} We follow previous works~\citep{lu2023mathvista, li2025eee} to first generate answers from models and subsequently using GPT-4o to extract the final answer from the answer content. For multiple-choice and short-form answers, we compare the extracted value directly against the ground-truth to calculate accuracy; for grounding type problems such as Jigsaw Game and Spot the Difference with list type answer, we calculate the intersection over union (IoU).
\paragraph{Implementation details.} We conduct all evaluations in zero-shot manner for fair comparison and better generalization. For open models, all experiments are done on NVIDIA A100 GPUs. For proprietary models, we use the official API. More details can be found in the \autoref{app:imple}. 

\begin{table}[t]
\centering
\caption{Model Accuracy (\%) Across Various Evaluation Tasks. SR: Symbolic Reasoning, WS: Word Search, LL-VQA: Low-Light VQA, IR: Instrument Reasoning, SD: Spot Difference, JG: Jigsaw Game, VS: Visual Search, RG: Rotation Game, Pro.: Proportion VQA. o3-TU: o3-tool-using, i.e., o3 with code interpreter. o3: o3 without code interpreter.}
\label{tab:model_accuracy}

\resizebox{\columnwidth}{!}{
\begin{tabular}{@{}l|c|ccccccccccccc@{}}
\toprule
\textbf{Model} & 
    All &
    Color & 
    Pro. & 
    OCR & 
    SR & 
    Maze & 
    Math & 
    WS & 
    LL-VQA & 
    IR & 
    SD & 
    JG & 
    VS & 
    RG \\ 
\midrule
Random Guess    & 13.5        & 28.0        & 6.7         & -           & 14.0        & 13.3        & 15.8        & 0.0         & 4.0         & 8.8         & 22.6        & 5.8         & 22.5        & 16.0 \\
\midrule
\multicolumn{15}{@{}c}{\textit{Open-Source MLLMs}} \\
\midrule
Llava-1.6-M-7B  & 11.3        & 27.0        & 7.5         & 3.3         & 16.0        & 4.2         & 16.7        & 0.0         & 14.0        & 6.3         & 18.0        & 0.0         & 22.5        & 12.0 \\
Llava-1.6-V-7B  & 11.5        & 27.0        & 10.8        & 0.0         & 10.0        & 15.0        & 12.5        & 1.0         & 8.0         & 7.5         & 12.2        & 0.0         & 24.2        & 13.3 \\
Llava-1.6-34B   & 13.0        & 31.0        & 6.7         & 1.7         & 20.0        & 15.8        & 18.3        & 0.0         & 16.0        & 16.3        & 11.9        & 0.0         & 21.7        & 10.7 \\
Llava-Next-72B  & 11.2        & 20.0        & 15.8        & 3.3         & 8.0         & 10.8        & 15.0        & 0.0         & 10.0        & 11.3        & 16.3        & 0.0         & 23.3        & 12.0 \\
Qwen2.5-VL-3B   & 17.7        & 27.0        & 20.0        & 31.7        & 12.0        & 21.7        & 20.8        & 0.0         & 12.0        & 13.8        & 29.7        & 0.0         & 26.7        & 12.0 \\
Qwen2.5-VL-7B   & 16.0        & 21.0        & 10.8        & 48.3        & 14.0        & 15.0        & 24.2        & 0.0         & 22.0        & 11.3        & 24.4        & 0.0         & 21.7        & 9.3  \\
Qwen2.5-VL-32B  & 18.7        & 26.0        & 19.2        & 25.0        & 10.0        & 18.3        & 23.3        & 2.0         & 14.0        & 15.0        & 13.1        & 5.4         & 48.3        & 13.3 \\
Qwen2.5-VL-72B  & 19.7        & 37.0        & 15.0        & 33.3        & 24.0        & 35.0        & 22.5        & 3.0         & 32.0        & 12.5        & 14.1        & 0.0         & 25.8        & 12.0 \\
InternVL3-8B    & 16.9        & 23.0        & 11.7        & 0.0         & 6.0         & 33.3        & 21.7        & 2.0         & 22.0        & 8.8         & 16.6        & 4.5         & 36.7        & 17.3 \\
InternVL3-38B   & 19.1        & 24.0        & 10.8        & 3.3         & 26.0        & 23.3        & 29.2        & 8.0         & 28.0        & 13.8        & 14.6        & 5.1         & 44.2        & 13.3 \\
InternVL3-78B   & 21.4        & 25.0        & 21.7        & 3.3         & 24.0        & 32.5        & 23.3        & 8.0         & 28.0        & 16.3        & 18.9        & 5.8         & 39.2        & 26.7 \\

\midrule
\multicolumn{15}{@{}c}{\textit{Proprietary MLLMs}} \\
\midrule
GPT-4.1         & 18.8        & 36.0        & 7.5         & 11.7        & 12.0        & 17.5        & 25.0        & 4.0         & 24.0        & 11.3        & 30.9        & 5.1         & 34.2        & 22.7 \\
GPT-4o          & 17.3        & 26.0        & 22.5        & 10.0        & 10.0        & 20.0        & 15.8        & 0.0         & 26.0        & 7.5         & 19.4        & 6.2         & 35.0        & 20.0 \\
Gemini-2.5-Flash         & 25.2        & 34.0        & 20.0        & 30.0        & 26.0        & 17.5        & 30.8        & 10.0        & 42.0        & 13.8        & 18.5        & 8.0         & 55.8        & 29.3 \\
Gemini-2.5-Pro           & 28.9        & 44.0        & 21.7        & 25.0        & 34.0        & 24.2        & 30.8        & 12.0        & 42.0        & 20.0        & 28.5        & 10.4        & \best{58.3} & 30.7 \\
Grok-4          & 22.5       & 35.0       & \best{53.3} & 6.7        & 20.0       & 25.8       & 19.2       & 2.00       & 22.0       & 12.5      & 27.0       & 10.0        & 25.8     & 18.7 \\

o4-mini             & 21.2        & 39.0        & 17.5        & 8.3         & 12.0        & 13.3        & 21.7        & 5.0         & 30.0        & 18.8        & 33.0        & 8.0         & 39.2        & 26.7 \\
o3              & 26.9        & 36.0        & 34.2 & 8.3         & 34.0        & 29.2        & 24.2        & 4.0         & 28.0        & 17.5        & 37.2        & 10.8        & 47.5        & 33.3 \\
\midrule
\multicolumn{15}{@{}c}{\textit{Tool-Using MLLMs}} \\
\midrule
DeepEyes & 17.3&22.0 & 6.7& 41.7& 19.9& 16.7& 20.0&1.0 & 16.0& 3.8&19.9&3.9&50.8&12.0\\
PyVision & 31.8&53.0 &26.7 &\best{63.3} &54.0 & 15.8&25.8 &10.0 &32.0 &17.5 &36.4&7.6&55.0&46.7\\
o4-mini-TU          & 37.5        & 53.0        & 21.7        & 53.3 & 58.0        & 34.2        & 31.7        & 55.0        & \best{44.0} & 13.8        & 38.9        & 11.8        & 47.5        & 52.0 \\
o3-TU           & \best{46.0} & \best{55.0} & 31.7        & 53.3 & \best{66.0} & \best{42.5} & \best{50.0} & \best{64.0} & 42.0        & \best{21.3} & \best{41.0} & \best{16.4} & 57.5        & \best{77.3} \\
\bottomrule
\end{tabular}%
} 
\end{table}

\subsection{Experimental Results}
\label{exp_results}
Both average and task-wise accuracies are reported in Table \ref{tab:model_accuracy}. We discuss several findings below. 

\paragraph{Result 1: TIR-Bench is challenging for all model types.} The highest performance observed among all models is only 46\%, a result that underscores the difficulty of the TIR-Bench. This benchmark proves to be a significant challenge for assessing Thinking-with-Images capabilities, even for advanced models such as o3-TU, which leverage a code interpreter.

\paragraph{Result 2: Traditional non-agentic models perform poorly on TIR-Bench.} Across all tasks, non-tool-using MLLMs show poor performance: the performance of most open-source models is close to or slightly higher than the  random guess performance while the top-performing MLLM, Gemini-2.5-pro, surpasses random guess results by only 15\%. These results highlight that, without agentic tool-using abilities, MLLMs can not perform well on TIR-Bench.

\paragraph{Result 3: Agentic tool-using is essential for TIR-Bench.} The o3-TU model \citep{OpenAI_o3_2025} demonstrates the strongest overall performance, achieving the highest average accuracy at 46\%. This represents a substantial lead, outperforming the Gemini-2.5-Pro model by nearly 17\% and the o3 model without a code interpreter by 19\%.  With tool-use enabled, it achieves state-of-the-art results on the majority of the tasks, winning 10 out of the total 14 categories. The o3-TU and o4-TU models demonstrate a large performance improvement on tasks involving straightforward image manipulation or processing, including the rotation game, rotated image OCR, and word search. Similar phenomenon appears in PyVision, which implements thinking-with-images abilities based on GPT-4.1. PyVision brings 13\% accuracy improvement compared with GPT-4.1.

We also observe that, although o3-TU excels in most categories, the improvements are not uniform. For example, for complex tasks such as \emph{Jigsaw game}, the performance of o3-TU is still very low. On \emph{ Proportion VQA}, performance surprisingly decreases from 34.2\% to 31.7\%. This task requires calling external segmentation models such as  Segment Anything~\citep{kirillov2023segment} to obtain a good rough estimate of the object segments. However, the current o3-TU model can only write executable code to manipulate images, but lacks the capability to call segmentation models specifically.

\begin{figure}[t]
  \centering
  \includegraphics[width=\linewidth]{ 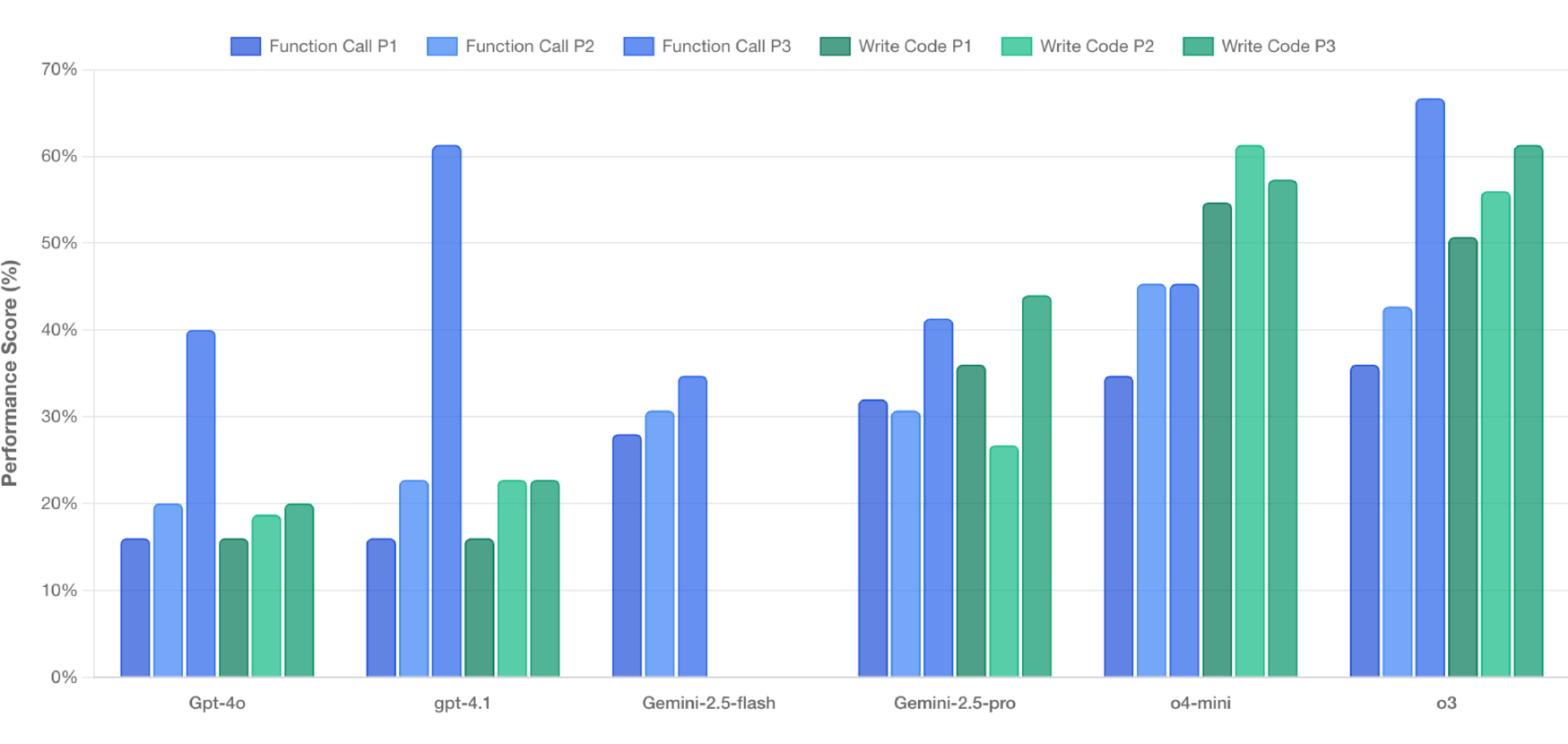}
  \caption{Comparison of function calling and writing code ability of different models.}
  \label{fig:f_call}
\end{figure}

\subsection{Qualitative Analysis}

\begin{figure}[t]
  \centering
  \includegraphics[width=\linewidth]{ 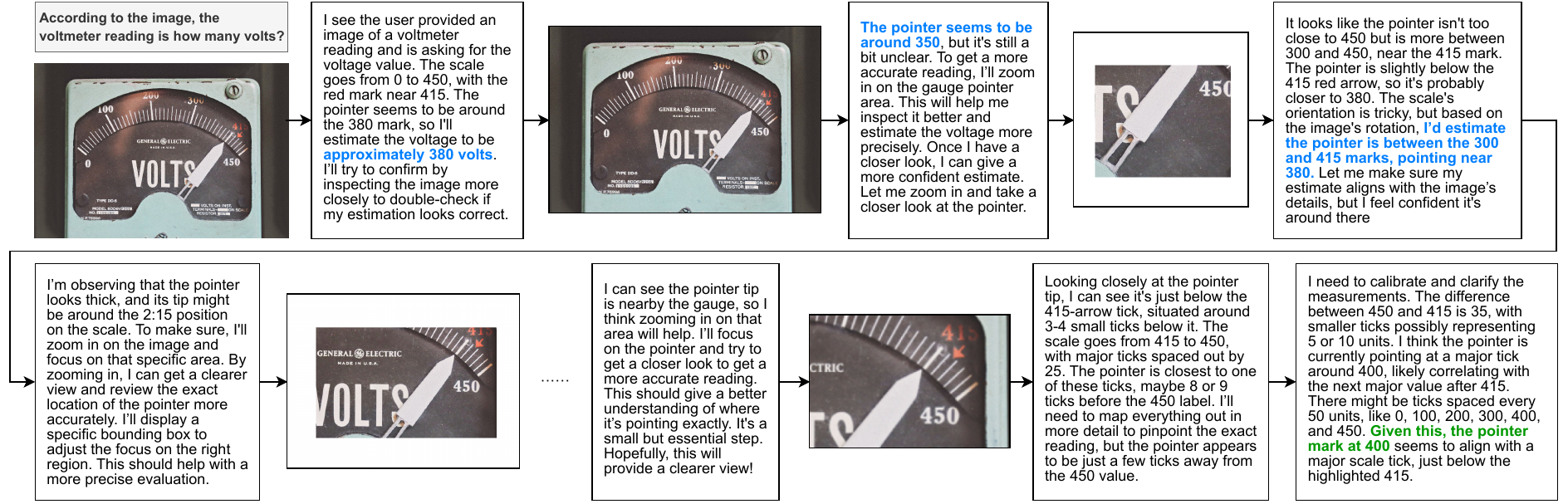}
  \caption{An example of o3 solving an case of instrument reading from TIR-Bench.}
  \label{fig:demo1}
\end{figure}

\begin{figure}[t]
  \centering
  \includegraphics[width=\linewidth]{ 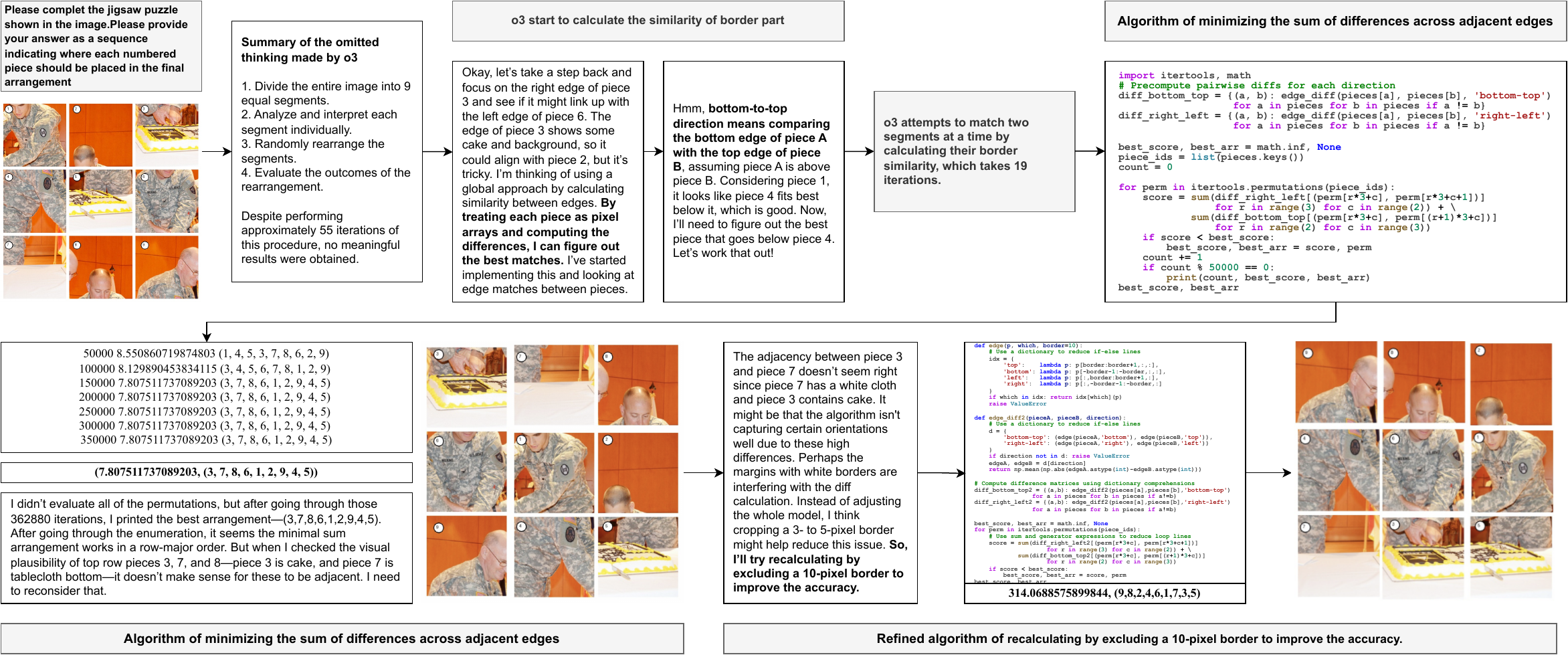}
  \caption{The o3 model’s reasoning process for a jigsaw puzzle task in TIR-Bench involved a large number of ineffective attempts.}
  \label{fig:demo2}
  \vspace{-1em}
\end{figure}

We present several examples of o3-TU responses here and examples of other model responses in Appendix.~\ref{app:example}. . Since the responses from the OpenAI API contain no reasoning process, we re-ran the questions in the web-based ChatGPT interface and analyzed the responses. Figure~\ref{fig:demo1} shows an case of instrument reading.
The model repeatedly crops the region around the pointer, checks its position, and guesses the answer (highlighted in \textcolor{myblue}{blue}). After several cycles of cropping, observing, and reflecting, o3 finally confirms its answer as ``400,'' which is correct.
Figure~\ref{fig:demo2} shows a case of jigsaw puzzle involved a lot of ineffective attempts. The model first divided the complete image into segments, then tried to compare and judge similarity by “understanding” the image as a whole. In the early empirical stage, the model made nearly 55 iterations, but failed to obtain any usable partial results. After that, it began to use edge comparison, sampling the borders of image pieces and comparing them pair by pair, which also did not produce effective outcomes. The model then started to develop algorithms, applying brute-force permutations to explore possible arrangements and calculate similarity. However, the initial algorithm did not work. After modifying the sampling method, the model, through more than 36,000 attempts, achieved a high similarity score that corresponded to the correct solution. This demonstrates that relying solely on the model’s raw visual capabilities is insufficient; only code-based perception proves to be reliable.
\subsection{Function Call Experiment Results}
\label{exp_funciton}
In this subsection, we report the results on the function calling ability of different MLLMs. 

\textbf{Set-up.} Using the Rotation Game task as a case study, we experimented with two distinct function-calling strategies: (1) providing the model with a predefined rotate function, requiring it only to output the degree parameter; and (2) requiring the model to generate the full image rotation code itself. Additionally, we tested three prompt variations to assess the impact of guidance: (1) a baseline prompt containing only the question; (2) a prompt including a hint to leverage the rotation function; and (3) a more explicit prompt instructing the model to systematically test each degree from the answer choices. The specific function definitions and prompts used in this analysis are provided in \autoref{appendix:prompt}. During inference, we repeatedly call functions until the model produces a final answer without requiring any function parameters.

\textbf{Results.} We report the experimental results in Figure.~\ref{fig:f_call} and report the average number of calling for each problem in Table.~\ref{tab:num_call}. Since Gemini-2.5-flash does not write code for all three prompts, we do not report it for writing code. We brief discuss some key findings here: \textbf{(1).} Clear Performance Hierarchy. o3 emerges as the top performer, achieving the highest accuracy, with o4-mini following closely.
\textbf{(2).} Prompting Strategy is Key. The prompt strategy hinting to check each degree choice (prompt 3) works best. The performance of most models increases with prompt 3 compared to the other two prompts. This implies that models may not know how to best utilize the functions without explicit guidance.
\textbf{(3).} Increased function call number in Newer Models. The average number of function calls for more recent advanced models (e.g., o3) is much higher than for previous models (e.g., GPT-4o). This implies that recent models are better trained for iterative function calling. 

\subsection{Fine-Tuning Comparison Experiment Results}
\label{exp_ft}
\begin{figure}[t]
  \centering 
  
  \begin{subfigure}[b]{0.48\textwidth}
    \centering
    \includegraphics[width=\textwidth]{ 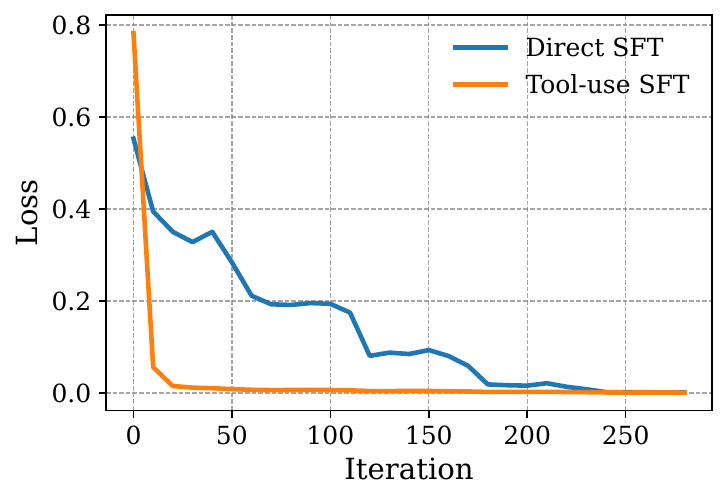} %
    \caption{Loss}
    \label{fig:first_image}
  \end{subfigure}
  \hfill 
  \begin{subfigure}[b]{0.48\textwidth}
    \centering
    \includegraphics[width=\textwidth]{ 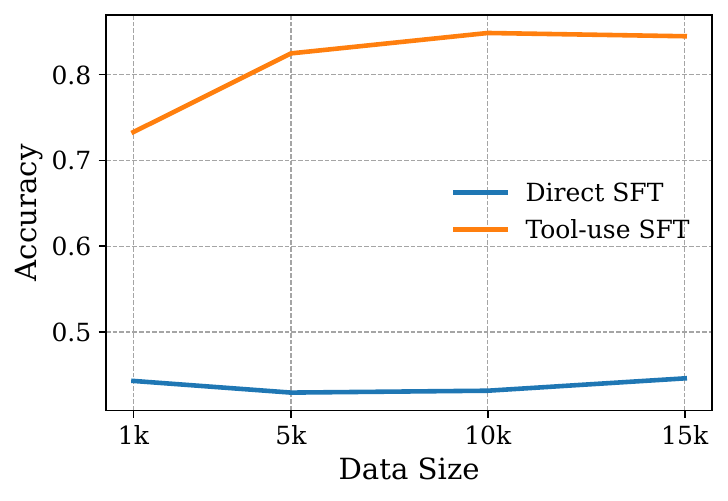} %
    \caption{Accuracy}
    \label{fig:second_image}
  \end{subfigure}
  
  \caption{Comparison of change of loss and accuracy between direct SFT and tool-use SFT.} 
  \label{fig:two_images_side_by_side} %
  \vspace{-1em}
\end{figure}
In this subsection, we report the experimental results on the different fine-tuning strategies.

\textbf{Setup.}  We use the rotated image OCR task as a case study to evaluate data scaling performance. We created training sets of varying sizes: 1k, 5k, 10k, and 15k samples—by randomly selecting and rotating images from the OCRDataset~\citep{ducto489_ocr_datasets}. We then compared two distinct training strategies:
(1). Direct SFT: A standard supervised fine-tuning approach where the model is trained to map the rotated image directly to the ground-truth text.
(2). Tool-Use SFT: An agentic approach where the model first learns to output the correct rotation degree. The restored image is then concatenated with the original context, and the model is subsequently trained to read the text from this corrected visual input. We use Qwen-2.5-VL-7B and fully fine-tune all parameters with 5 epochs. 

\textbf{Results.} We report accuracy on the Rotated OCR task in Figure~\ref{fig:second_image} and the loss curves on 15k samples in Figure~\ref{fig:first_image}. Overall, Tool-use SFT significantly outperforms Direct SFT. For Tool-use SFT, performance scales positively with data size, whereas Direct SFT shows no such trend. This suggests that simply scaling data for Direct SFT is ineffective on tasks requiring image-based reasoning. We also observe that Tool-use SFT’s loss decreases much faster despite starting from a higher initial value. This is likely because Qwen-2.5-VL was not pretrained with function-calling, leading to higher initial loss. Furthermore, since Qwen-2.5-VL was trained mostly on correctly oriented OCR data, fine-tuning it directly on rotated data may cause forgetting. In contrast, restoring image orientation before text output avoids this issue, which may explain the faster loss reduction.

\section{Conclusion}
In this paper, we propose TIR-Bench, a comprehensive benchmark designed to evaluate the thinking-with-images ability of agentic MLLMs. TIR-Bench is composed of 13 meticulously collected tasks that assess a wide range of tool-assisted reasoning skills. By assessing diverse MLLMs on TIR-Bench, including both standard models and those augmented with tool-
use capabilities, we find that TIR-Bench is a challenging benchmark
for all models that necessitates thinking-with-images capabilities for successful
completion. Lastly, we conduct a pilot study comparing direct and agentic fine-tuning for image-operation tasks and evaluating the function-calling abilities of various MLLMs.
\bibliography{iclr2026_conference}
\bibliographystyle{iclr2026_conference}

\appendix
\newpage
\section{The Use of Large Language Models~(LLMs)}
In this paper, we only used LLMs to correct the grammar and spelling errors in the writing. All the results are produced by authors.

\begin{figure}[t]
  \centering
  \includegraphics[width=\linewidth]{ 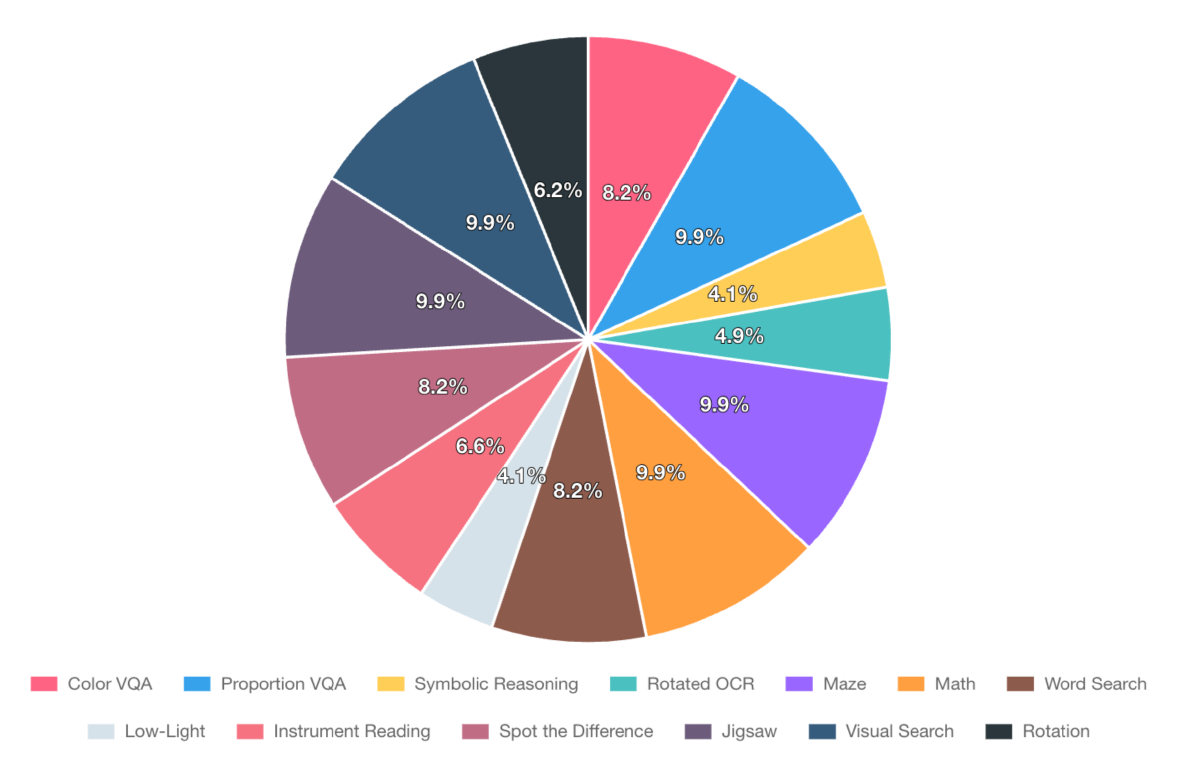}
  \caption{Task Distribution of TIR-Bench.}
  \label{fig:task_dis}
\end{figure}
\section{Detailed Task Design and Collection}
\label{appendix:task}
\subsection{Task Design}
\label{app:task_design}
To extensively validate the model's ability to think with images we design 13 tasks. The overview of the benchmark is shown in Figure \ref{fig:demo}. We describe these 13 tasks below:
\begin{itemize}[leftmargin=10pt]
    \item \textbf{Color VQA task:} This task assesses the model's ability to answer questions related to an image's color composition. Answering these questions requires the model to programmatically process the image to obtain visual information—for instance, by writing and executing code to calculate the proportion of a specific color. 
    \item \textbf{Referring Object Proportion VQA}: This task assesses the model's agentic capabilities, requiring it to call a powerful external segmentation model to obtain an object's mask and then programmatically calculate its proportion relative to the entire image.
    \item \textbf{Rotated Image OCR}: This task evaluates a model's ability to execute a multi-step visual reasoning process. The model must first identify that an image containing text is incorrectly oriented, then use a tool to rotate the image to its correct position, and finally perform optical character recognition (OCR) to accurately read the content.
    \item \textbf{Symbolic Reasoning}: This task assesses the model's ability to apply abstract, rule-based logic to visual information. For instance, when asked to count the edges of a complex polygon, the model cannot simply guess; it must systematically identify and enumerate each distinct edge, a process that may require internal algorithms for vertex or line detection to arrive at the correct count. 
    \item \textbf{Maze}: This task assesses the model's ability for advanced spatial planning and algorithmic execution. The model must analyze the visual structure of the maze, devise a solution using image processing tools (such as morphological operations), and apply a pathfinding algorithm to solve it. Finally, it must draw the solution path onto the image, demonstrating its ability to translate an abstract plan into a concrete visual action. 
    \item \textbf{Math Problems}: This task assesses the model's ability to solve geometric problems by programmatically augmenting the visual input. To find a solution, the model must use tools to draw auxiliary constructs, such as adding lines to a diagram or imposing a coordinate system to define and calculate properties like relative lengths. 
    \item \textbf{Word Search Puzzle}: This task evaluates the model's ability to perform fine-grained visual discrimination and anomaly detection. The image presents a field of numerous, nearly identical characters, with only a few subtle differences. Standard OCR is designed to fail in this scenario, forcing the model to devise a programmatic solution. To succeed, the model must write and execute code to perform a more fundamental analysis, such as a pixel-level comparison or a targeted visual search, to locate and identify the characters that deviate from the pattern. 
    \item \textbf{Low-Light Image VQA}: This task evaluates the model's ability to overcome suboptimal visual conditions. Given a low-light image where details are obscured, the model must first recognize the issue and then programmatically enhance the image, for instance, by writing and executing code to increase its contrast or brightness, before it can accurately answer questions about the content.
    \item  \textbf{Instrument Reading}: This task evaluates the model's ability to perform a sequential, tool-assisted analysis. To succeed, the model must first locate the key areas for reading the instrument, then programmatically crop and enlarge these specific regions for clarity, and finally, accurately read the value from the enhanced view. 
    \item \textbf{Spot the Difference}: This task evaluates the model's ability to perform precise, programmatic visual comparison. To identify the differences between two nearly identical images, the model must execute a tool-based strategy, such as programmatic image subtraction, to highlight differing regions at a pixel level. After locating the discrepancies, the model must then isolate and identify the specific image patches that contain these alterations.
    \item \textbf{Jigsaw Puzzle}: This task assesses the model's ability to perform complex spatial reasoning through an iterative, tool-based approach. The model is required to programmatically segment an image into pieces and then repeatedly attempt to reassemble them. This process involves a continuous loop of action and self-correction, where the model must evaluate each configuration to determine if the image has been successfully restored, continuing the cycle until the solution is achieved.
    \item \textbf{Visual Search}: This task evaluates the model's ability to locate specific targets within a complex or high-resolution image through deep, multi-turn reasoning. To succeed, the model must engage in an iterative search process, strategically using tools to repeatedly zoom in on and analyze different regions until the object or information is found. 
    \item \textbf{Rotation Game}: This task assesses the model's ability to perform iterative orientation correction~\citep{xie2025play}. To restore a rotated image to its upright position, the model must programmatically test various rotation angles. After each transformation, it must visually evaluate the result to determine if the orientation is correct, engaging in a trial-and-error loop of tool-based action and visual judgment until the problem is solved.
\end{itemize}

\section{Detailed Task Example}
\label{appendix:task_example}

\subsection{Color}
\begin{figure}[H]
    \centering
    \includegraphics[width=0.9\linewidth]{ 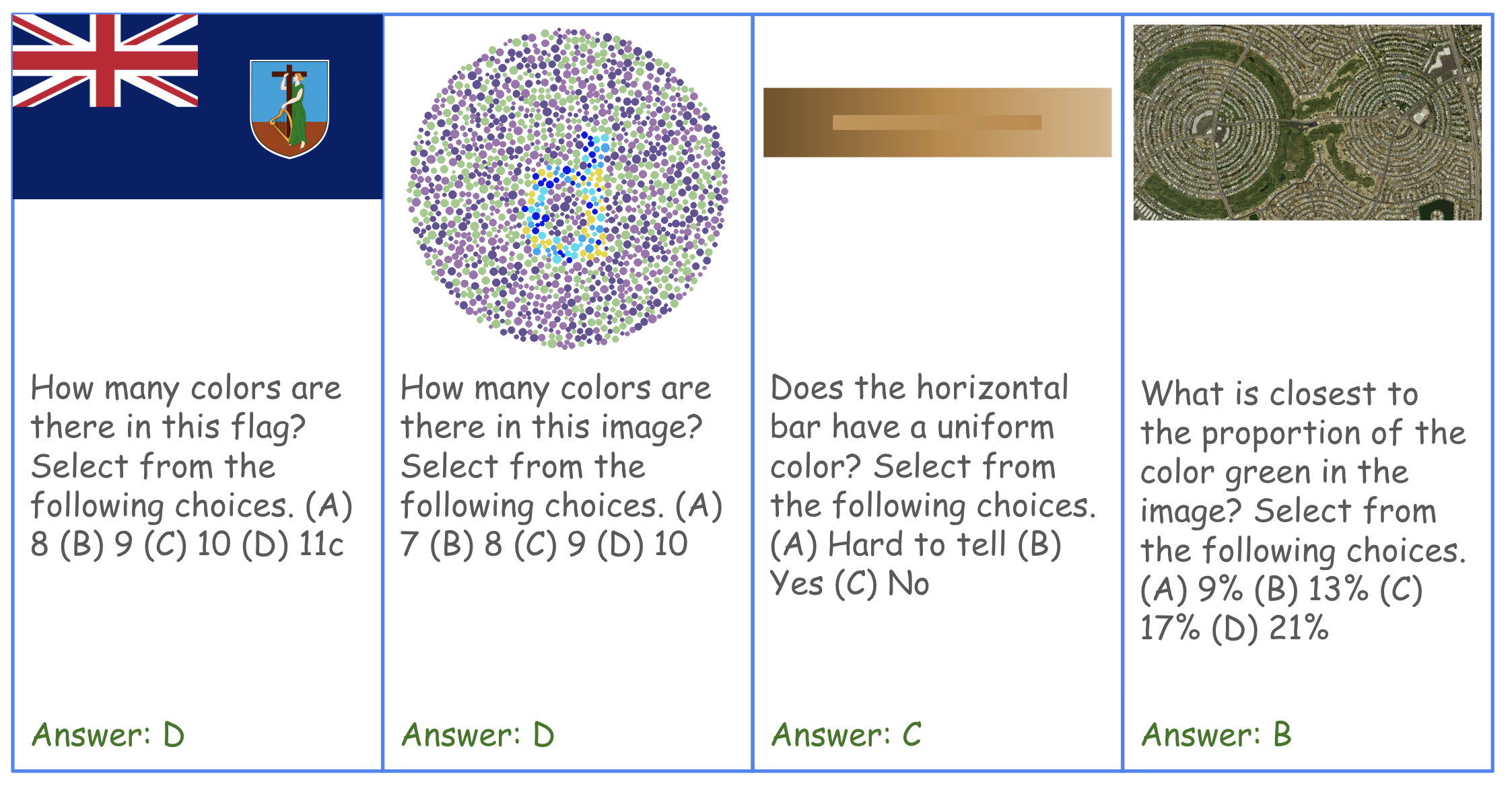}
    \caption{Additional example of ``Color'' task.}
\end{figure}

\subsection{Low-Light}
\begin{figure}[H]
    \centering
    \includegraphics[width=0.9\linewidth]{ 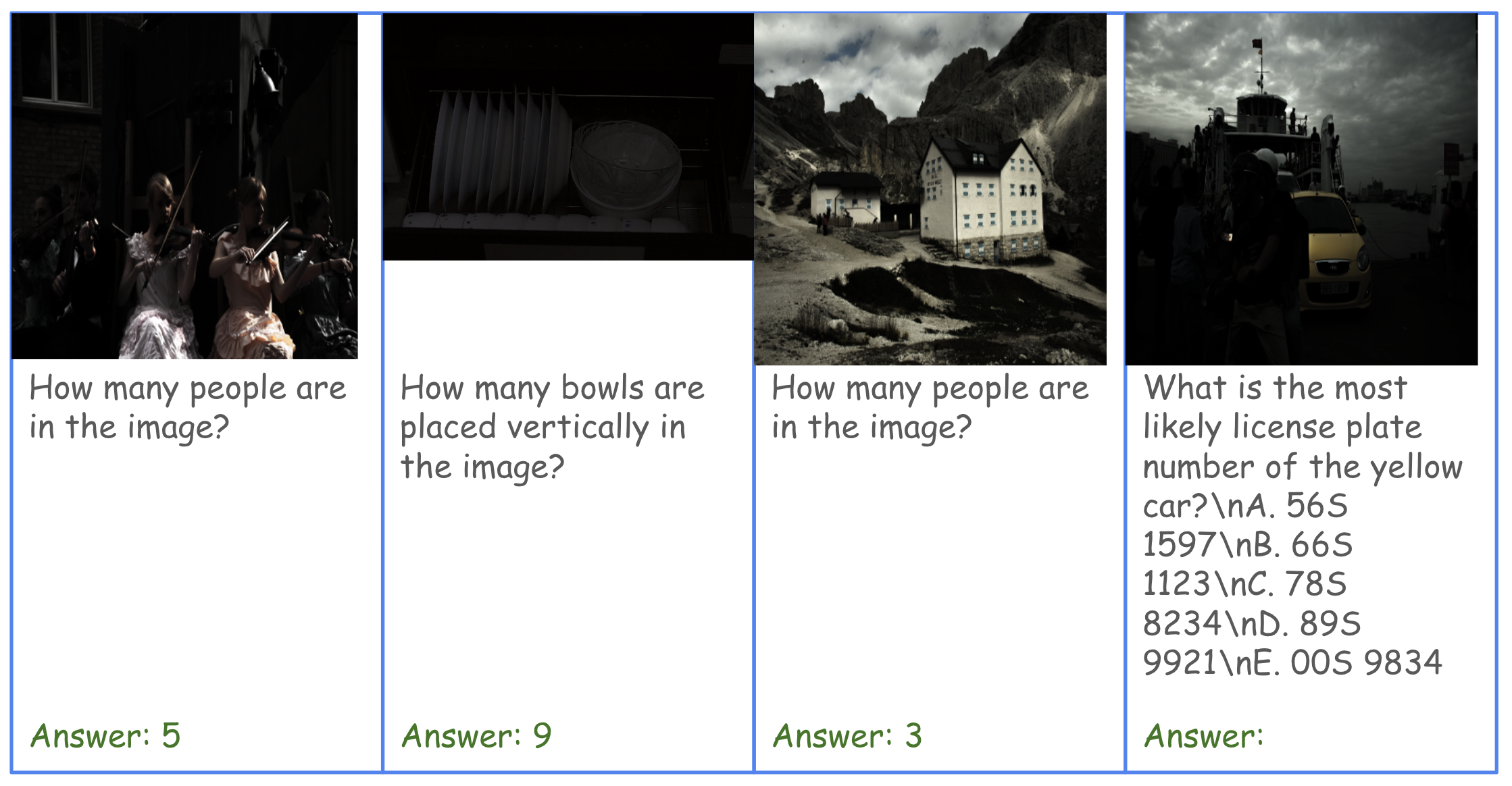}
    \caption{Additional example of ``Low-Light'' task.}
\end{figure}

\subsection{Instrument Reading}
\begin{figure}[H]
    \centering
    \includegraphics[width=0.9\linewidth]{ 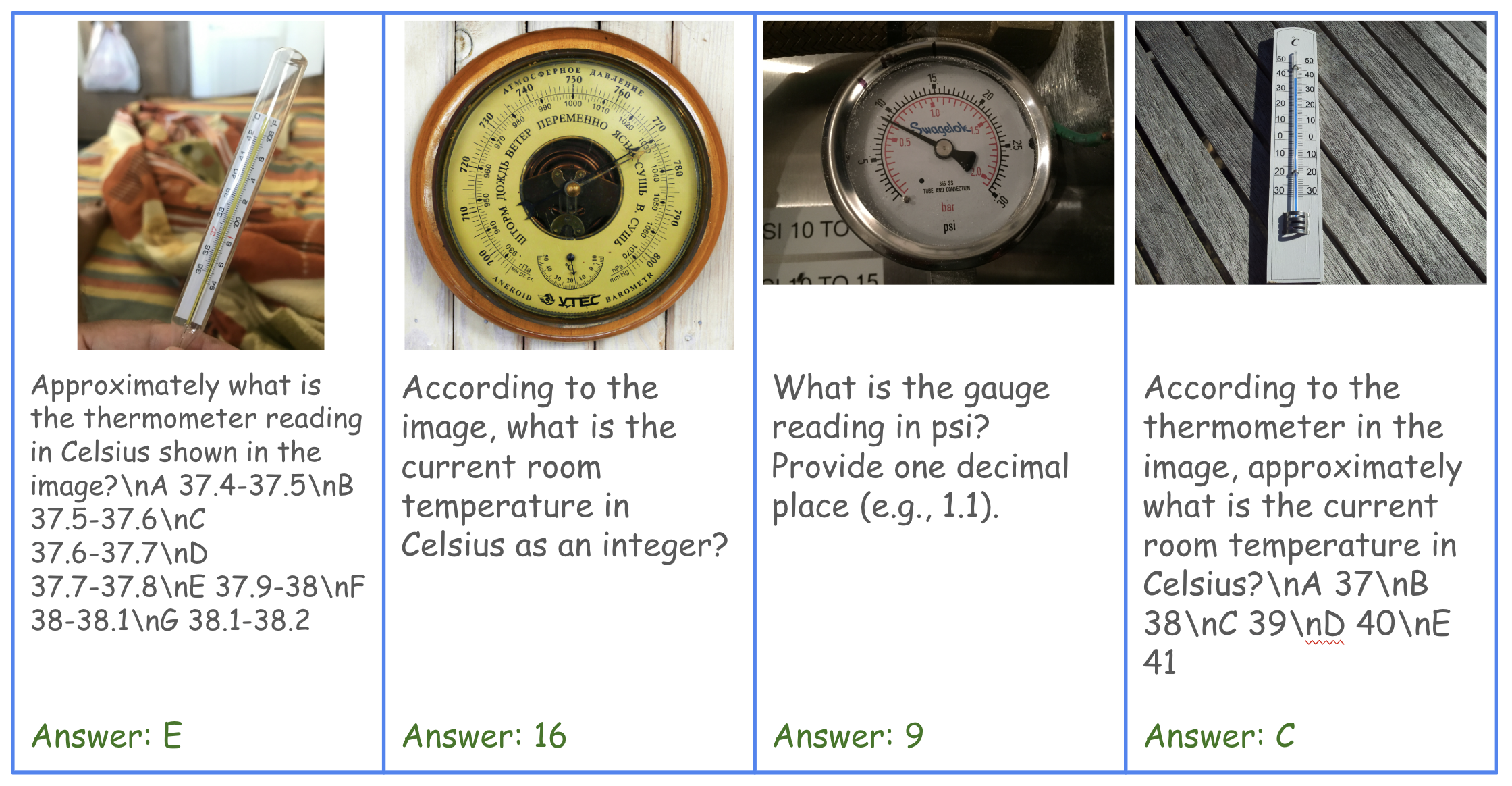}
    \caption{Additional example of ``Instrument Reading'' task.}
\end{figure}

\subsection{Jigsaw}
\begin{figure}[H]
    \centering
    \includegraphics[width=0.9\linewidth]{ 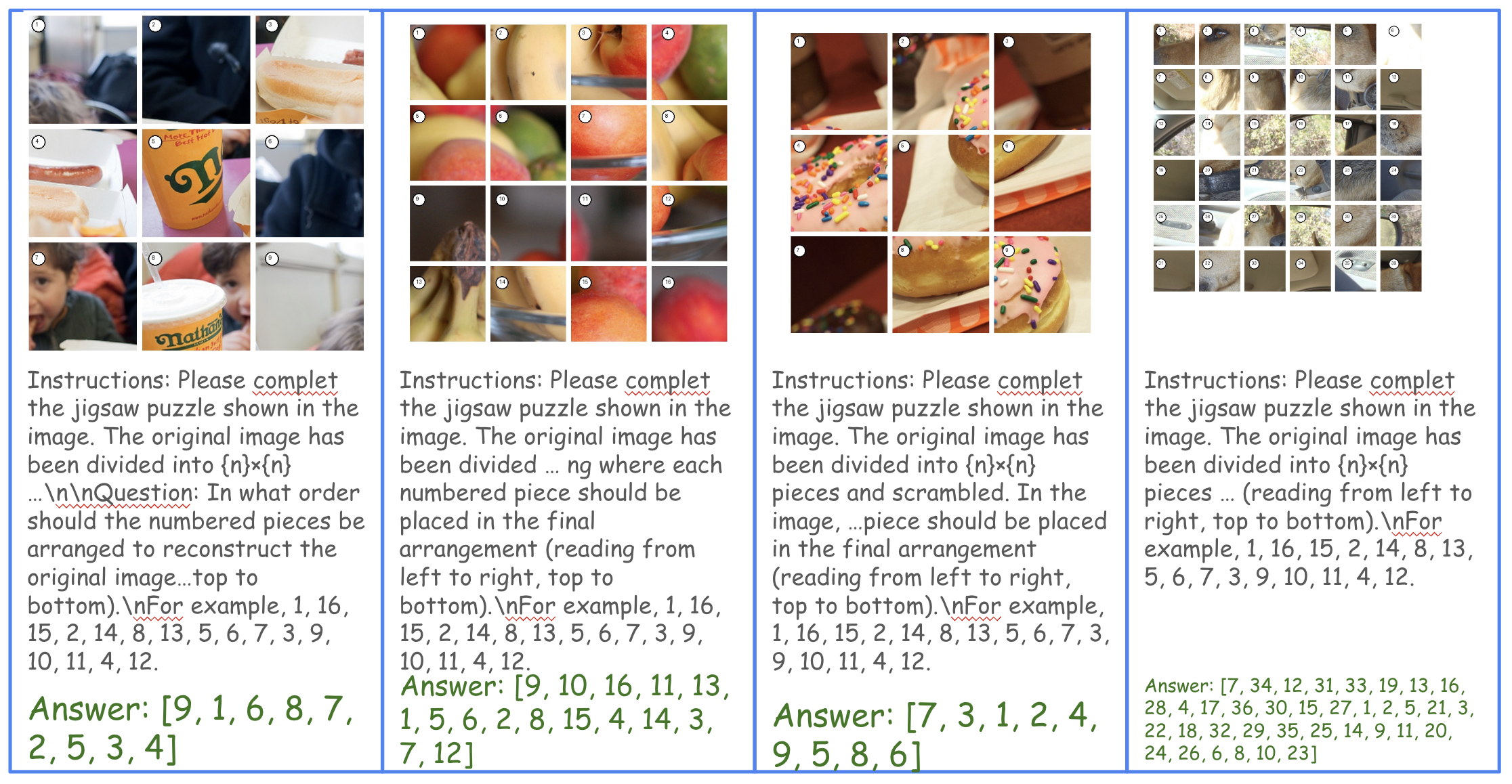}
    \caption{Additional example of ``Jigsaw'' task.}
\end{figure}

\subsection{Math}
\begin{figure}[H]
    \centering
    \includegraphics[width=0.9\linewidth]{ 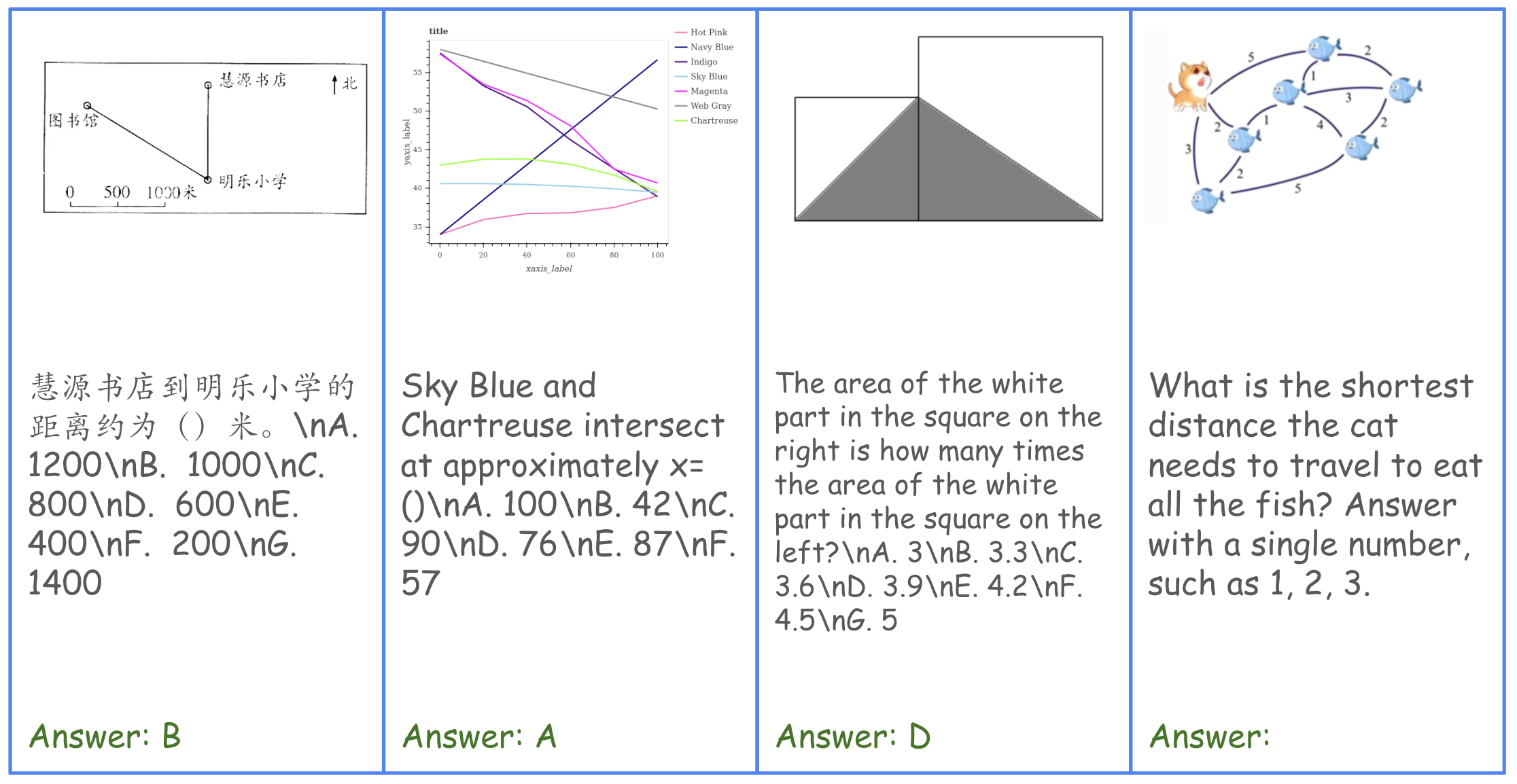}
    \caption{Additional example of ``Math'' task.}
\end{figure}

\subsection{Maze}
\begin{figure}[H]
    \centering
    \includegraphics[width=0.9\linewidth]{ 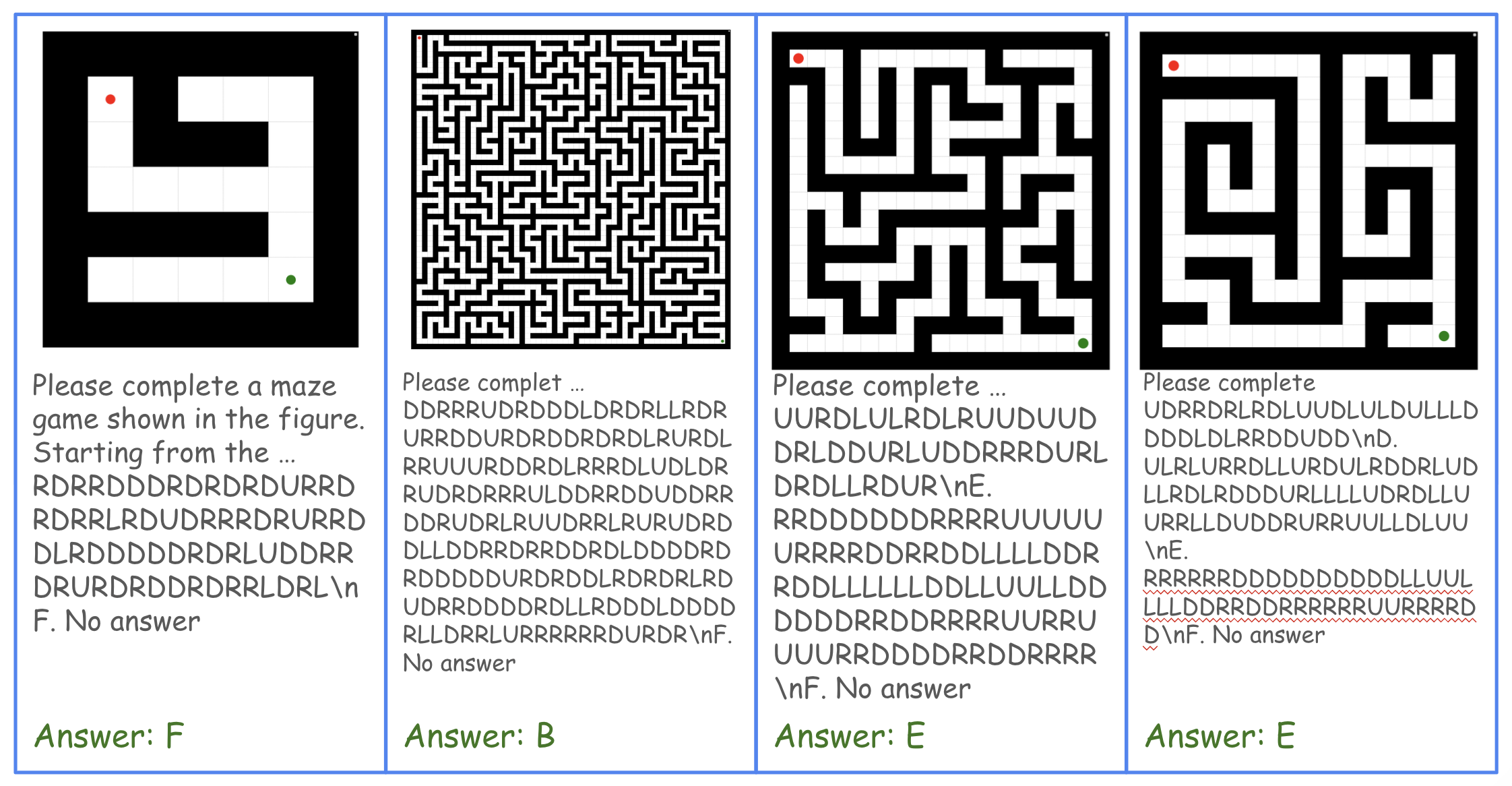}
    \caption{Additional example of ``Maze'' task.}
\end{figure}

\subsection{Rotated OCR}
\begin{figure}[H]
    \centering
    \includegraphics[width=0.9\linewidth]{ 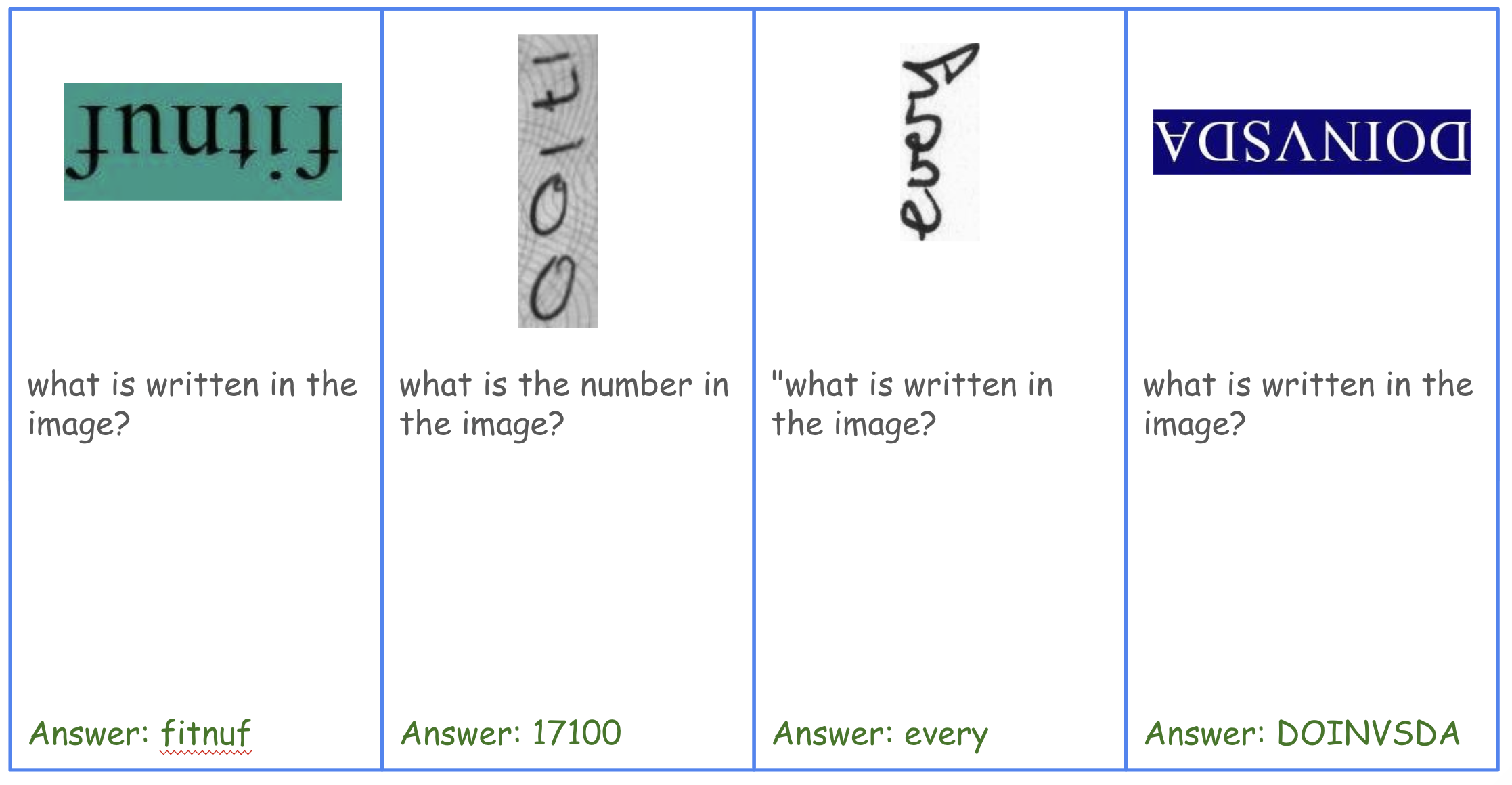}
    \caption{Additional example of ``Rotated OCR'' task.}
\end{figure}

\subsection{Proportion}
\begin{figure}[H]
    \centering
    \includegraphics[width=0.9\linewidth]{ 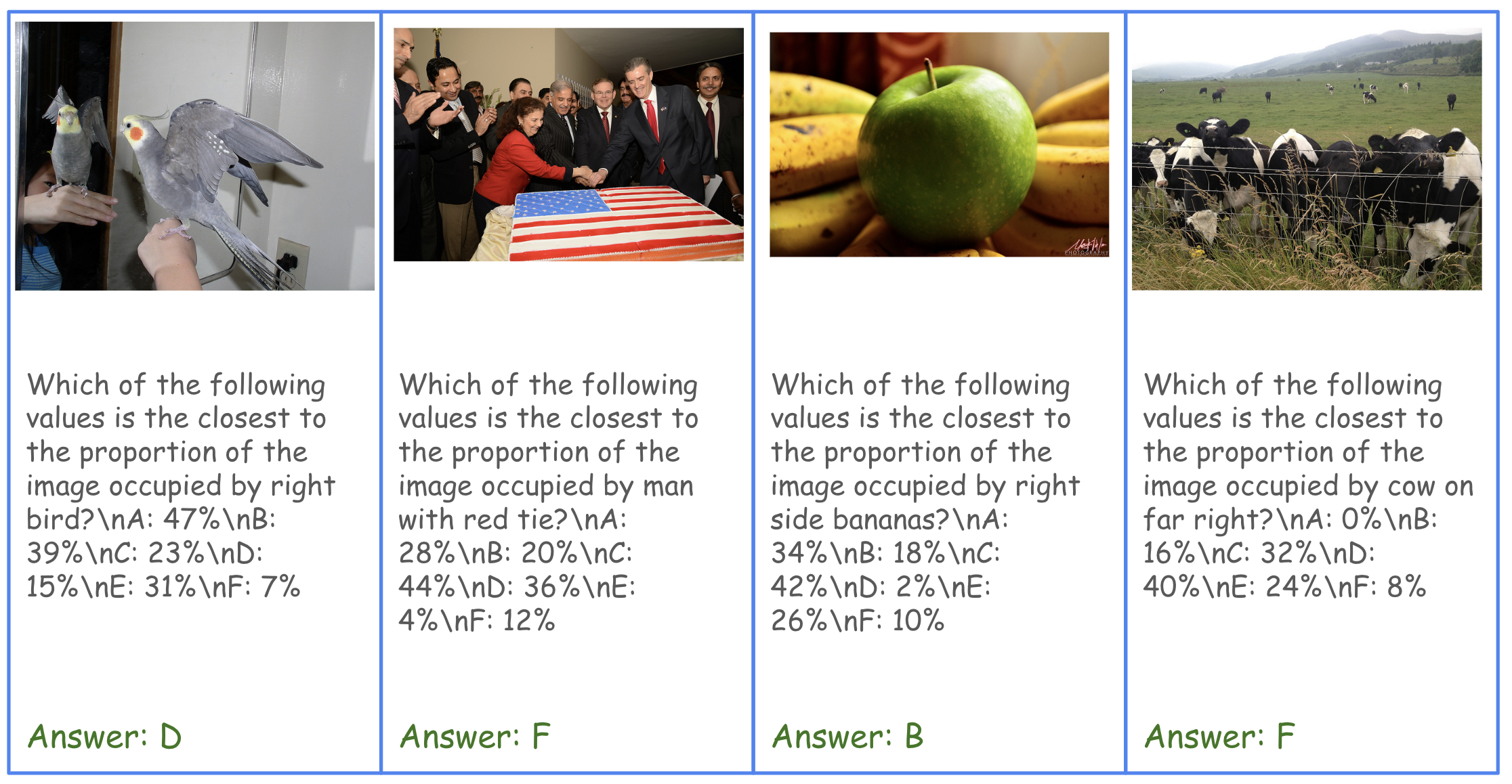}
    \caption{Additional example of ``Proportion'' task.}
\end{figure}

\subsection{Rotation}
\begin{figure}[H]
    \centering
    \includegraphics[width=0.9\linewidth]{ 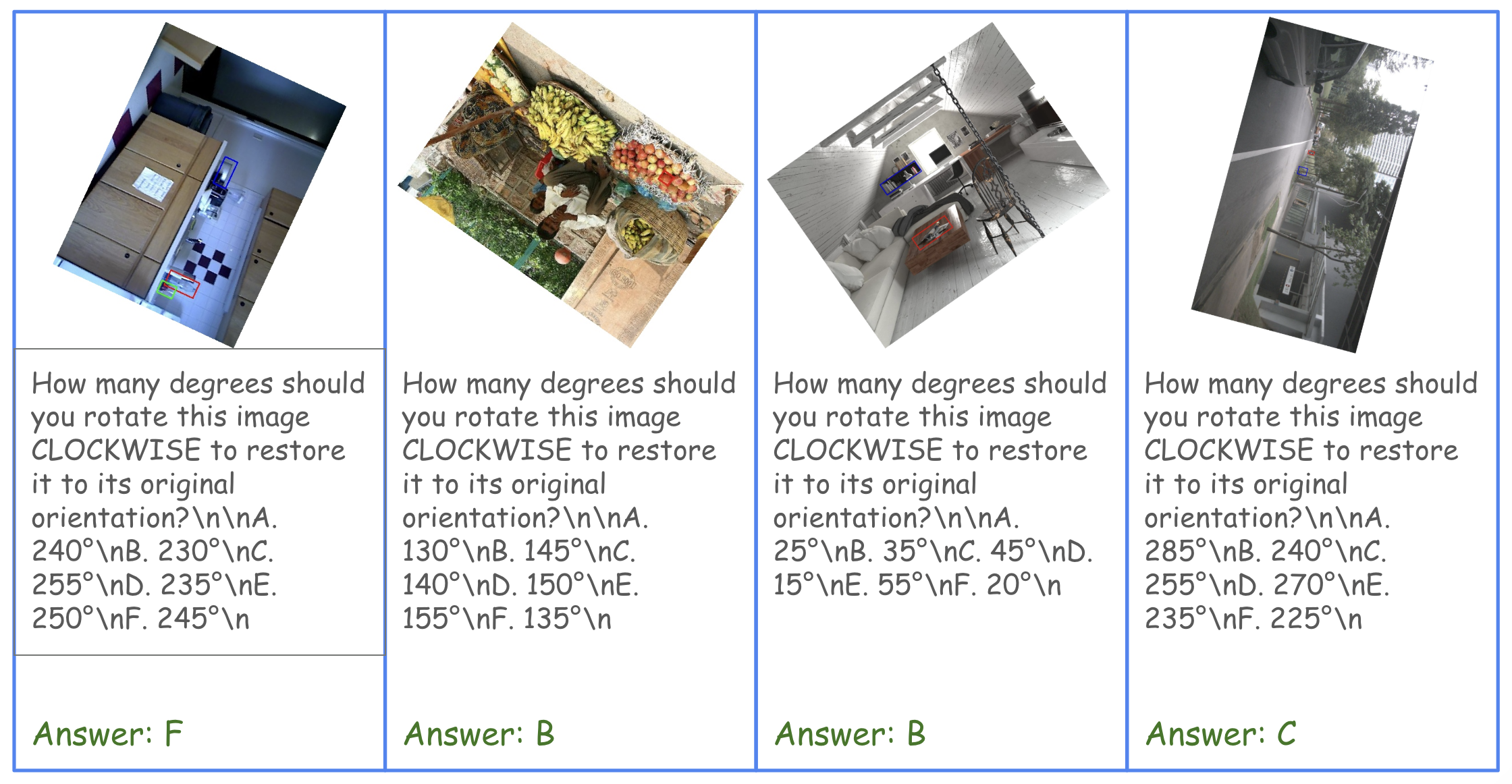}
    \caption{Additional example of ``Rotation'' task.}
\end{figure}

\subsection{Spot the Difference}
\begin{figure}[H]
    \centering
    \includegraphics[width=0.9\linewidth]{ 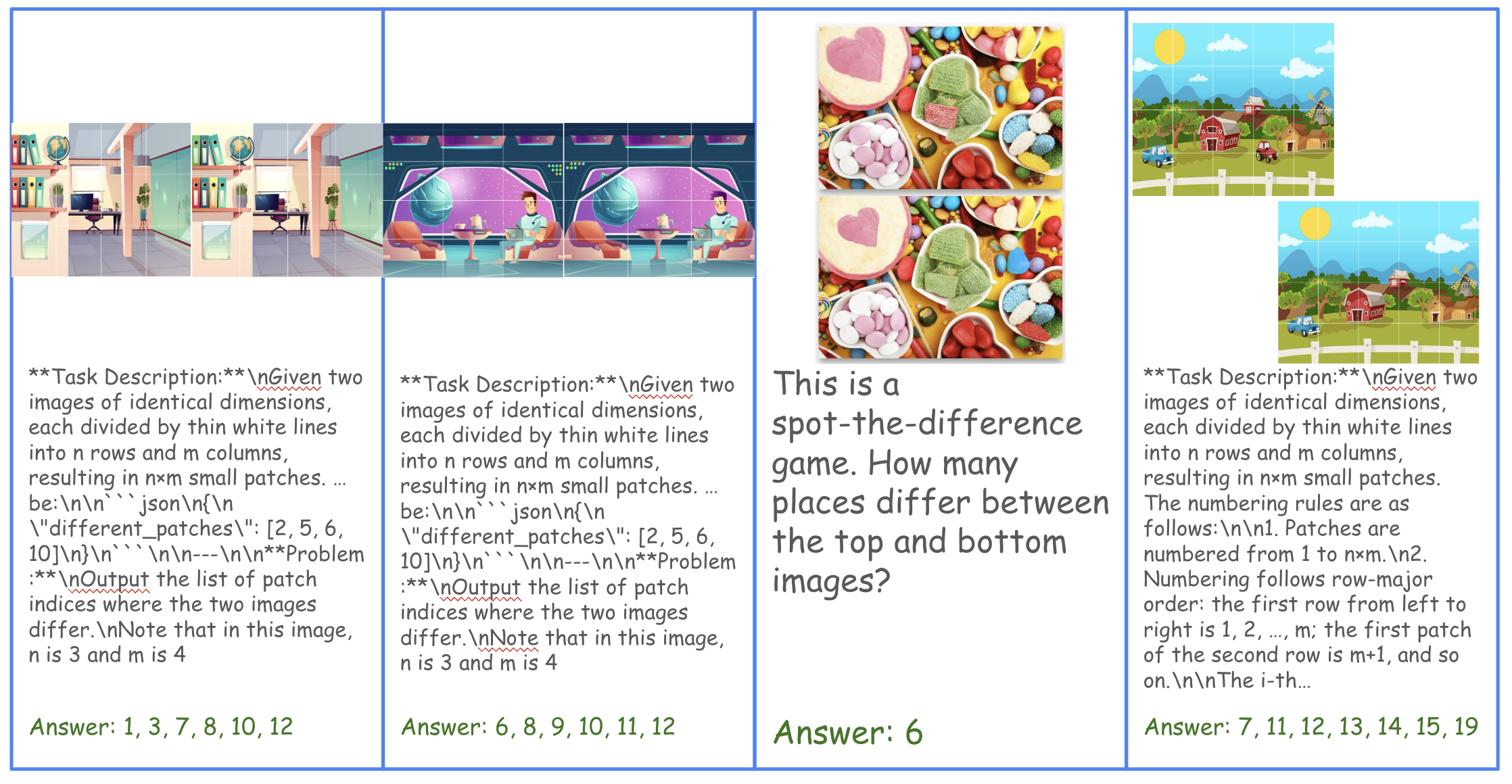}
    \caption{Additional example of ``Spot the Difference'' task.}
\end{figure}

\subsection{Symbolic Reasoning}
\begin{figure}[H]
    \centering
    \includegraphics[width=0.9\linewidth]{ 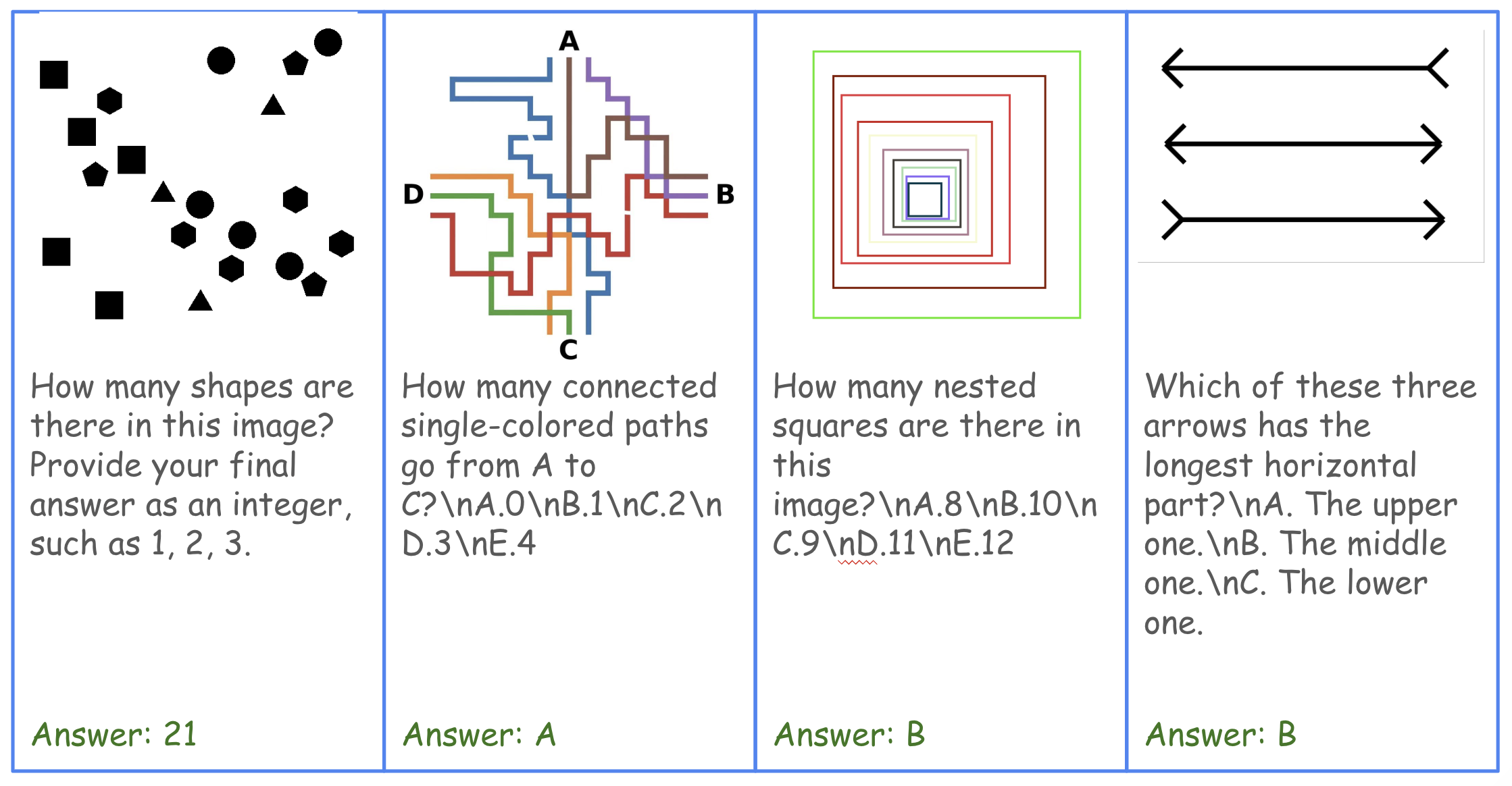}
    \caption{Additional example of ``Symbolic Reasoning'' task.}
\end{figure}

\subsection{Visual Search}
\begin{figure}[H]
    \centering
    \includegraphics[width=0.9\linewidth]{ 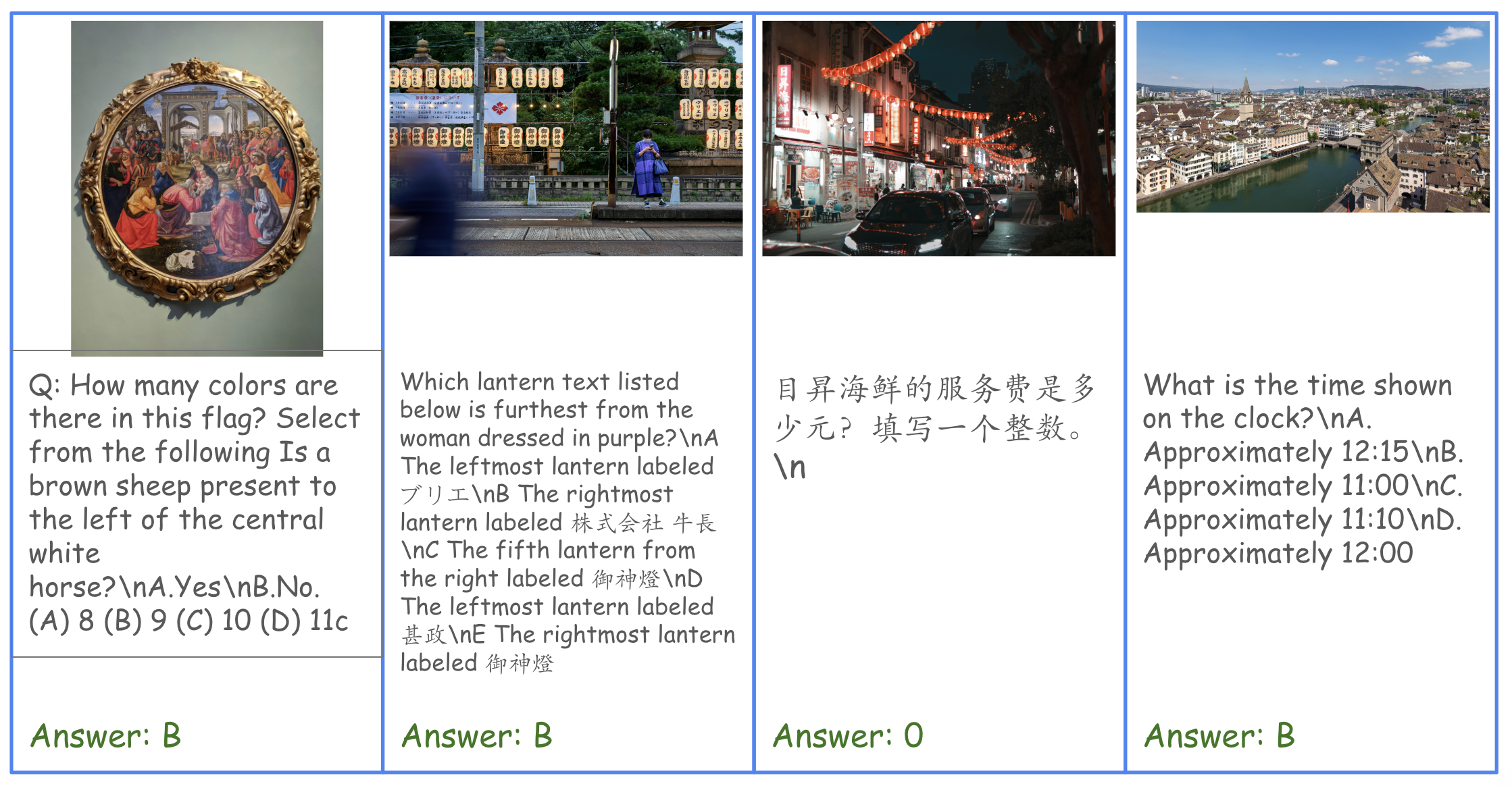}
    \caption{Additional example of ``Visual Search'' task.}
\end{figure}

\subsection{Word Search}
\begin{figure}[H]
    \centering
    \includegraphics[width=0.9\linewidth]{ 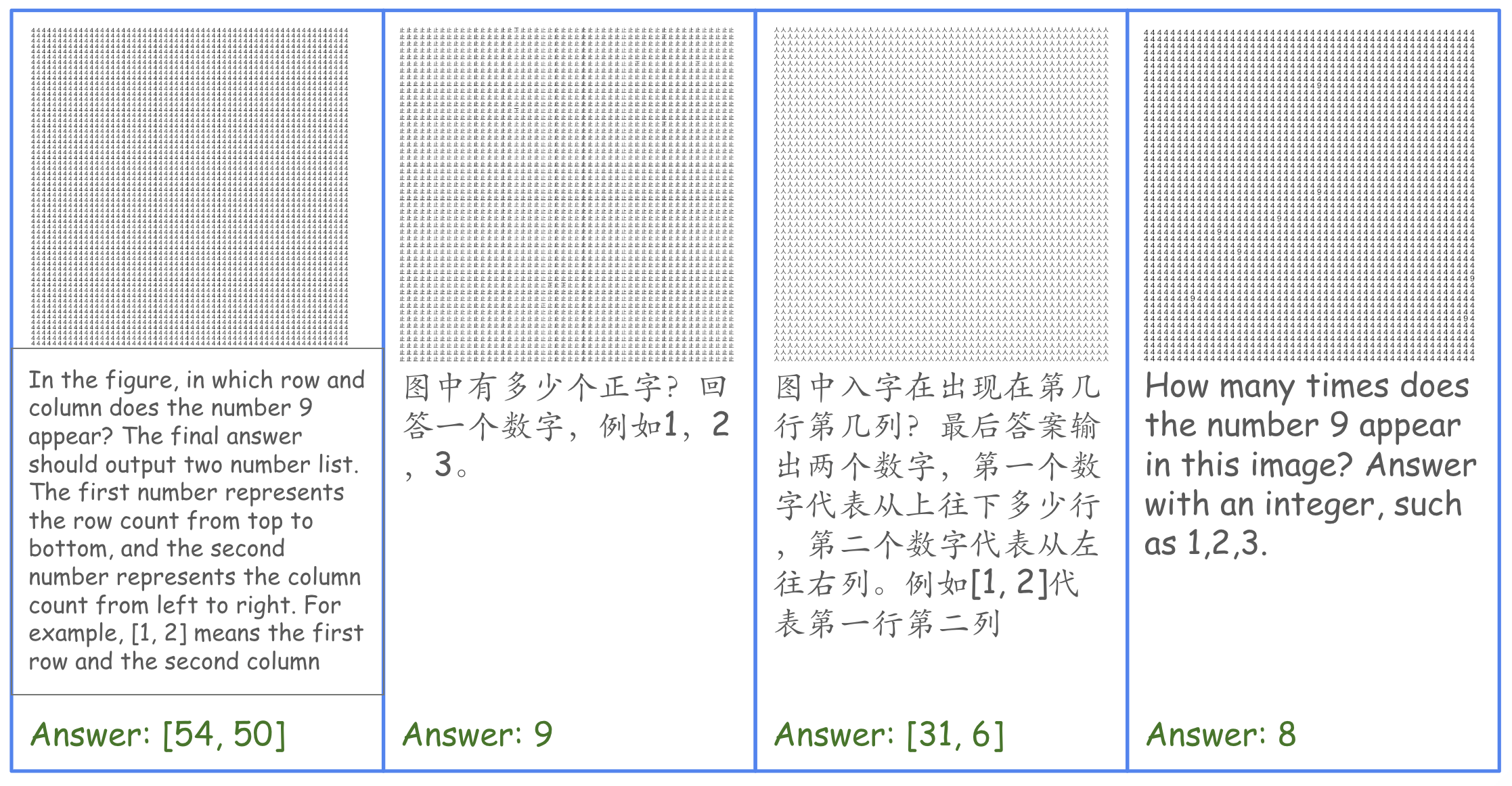}
    \caption{Additional example of ``Word Search'' task.}
\end{figure}

\section{Detailed Prompts}
\label{appendix:prompt}
\subsection{Write Code}
\texttt{prompt\_being} is the prompt for the beginning turn and \texttt{prompt\_return} is the prompt for returning processed images to models.

Prompt strategy 1:

\textit{%
\texttt{%
prompt\_1 = user\_prompt\\
prompt\_2 = "The returning status and the processed image or text (if any) of the code is attached."
}%
}

Prompt strategy 2:

\textit{%
\texttt{prompt\_being = user\_prompt + '\textbackslash nPlease consider to write code to process the image.'\\
prompt\_return = 'The returning status and the processed image or text (if any) of the code is attached, you can continue to write code to process the original image for better understanding or proceed to answer the question.'}%
}

For prompt 3:

\textit{%
\texttt{%
prompt\_1 = user\_prompt + "\textbackslash nPlease try to write code to rotate the original image with the rotation degree in the options to verify the correctness. You can try as many as you can."\\[2mm]
prompt\_2 = "The returning status and the processed image or text (if any) of the code is attached, you can continue to write code to rotate the original image with a degree of other difference options for better understanding or proceed to answer the question if you have meet the correct option."
}%
}
\subsection{Function calling}
For prompt strategy 1:

\textit{%
\texttt{%
prompt\_1 = user\_prompt\\
prompt\_2 = "The rotated image is attached."
}%
}

For prompt strategy 2:

\textit{%
\texttt{%
prompt\_1 = user\_prompt + '\textbackslash nPlease consider to call the rotation function.'\\
prompt\_2 = 'The rotated image is attached, you can continue to call the function to rotate the original image with a difference degree for better understanding or proceed to answer the question.'
}%
}

For prompt strategy 3:

\textit{%
\texttt{%
prompt\_1 = user\_prompt + "\textbackslash nPlease try to call the rotation function with the rotation degree in the options to verify the correctness. You can try as many as you can."\\
prompt\_2 = "The rotated image is attached, you can continue to call the function to rotate the original image with a degree of other difference options for better understanding or proceed to answer the question if you have meet the correct option."
}%
}
\begin{table}[ht]
\centering
\caption{Average number of function calling across different prompt strategies and calling stratefies.}
\label{tab:num_call}
\begin{tabular}{lcccccc}
\toprule
\multirow{2}{*}{Model} & \multicolumn{3}{c}{Function} & \multicolumn{3}{c}{Write Code} \\
\cmidrule(lr){2-4} \cmidrule(lr){5-7}
 & Prompt 1 & Prompt 2 & Prompt 3 & Prompt 1 & Prompt 2 & Prompt 3 \\
\midrule
GPT-4.1 & 0.00 & 1.15 & 4.68 & 1.32 & 1.68 & 3.95 \\
GPT-4o & 0.00 & 1.12 & 4.29 & 2.15 & 1.87 & 4.07 \\
Gemini-2.5-flash & 0.64 & 1.05 & 1.73 & - & - & - \\
Gemini-2.5-pro & 0.75 & 0.87 & 1.09 & 0.25 & 0.28 & 0.21 \\
o4-mini & 0.07 & 1.52 & 2.87 & 3.43 & 3.68 & 2.68 \\
o3 & 0.33 & 1.57 & 3.59 & 6.05 & 5.77 & 5.76 \\
\bottomrule
\end{tabular}
\end{table}

\section{Implementation Details}
We conduct all evaluations in zero-shot manner for fair comparison and better generalization. For open models, all experiments are done on NVIDIA A100 GPUs. For proprietary models, we use the official API. We set do sample = False, temperature = 0, max new tokens = 2048 for all open models.
\label{app:imple}

\section{Examples of Model Responses}
\label{app:example}
\begin{figure}[!ht]
\begin{tcolorbox}[colback=gray!10]
\begin{minipage}[t]{0.99\textwidth}
\includegraphics[width=0.5\textwidth]{ 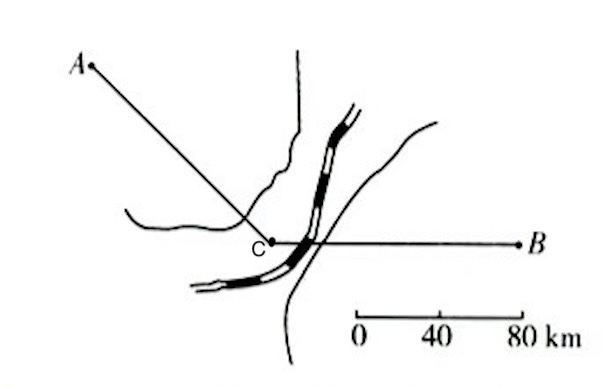}

\textbf{Question: } Measure the distances of AC and CB according to the scale. Driving from point A to point C and then from point C to point B, with a speed of 60 km/h, how many hours are needed? Answer with a single number, such as 1, 2, 3.

\textbf{Ground Truth: 4.}

\textbf{Response from o3-TU:} 4

\textbf{Response from o3: 3}.

\textbf{Response from o4-mini-TU:} 3

\textbf{Response from o4-mini:} 2

\textbf{Response from Gemini-2.5-Pro: }By measuring the distances using the provided scale, we find that the line segment CB is equal in length to the scale bar representing 80 km, so $\text{Distance CB} = 80$ km. The segment AC is shorter than the full 80 km scale; it appears to be three-quarters of the scale, giving $\text{Distance AC} = 0.75 \times 80 = 60$ km, or equivalently, 1.5 times the 0--40 km portion of the scale, which also gives 60 km. Therefore, the total distance is $\text{Total Distance} = \text{Distance AC} + \text{Distance CB} = 60 + 80 = 140$ km. Assuming a driving speed of 60 km/h, the time needed is $\text{Time} = \frac{\text{Total Distance}}{\text{Speed}} = \frac{140}{60} \approx 2.33$ hours, which rounded to the nearest whole number gives 2 hours.

\textbf{Response from Gemini-2.5-flash:} To solve this problem, we first determine the map scale: the scale bar shows that 223 pixels correspond to 80 km, so $1 \text{ pixel} \approx 0.3587 \text{ km}$. The pixel coordinates are $A(112,102)$, $C(352,417)$, and $B(686,417)$. Using the distance formula, the pixel length of $AC$ is $\sqrt{(352-112)^2+(417-102)^2} \approx 396.01$ pixels, and $CB$ is $\sqrt{(686-352)^2+(417-417)^2} = 334$ pixels. Converting to kilometers, $AC = 396.01 \times \tfrac{80}{223} \approx 142.06 \text{ km}$ and $CB = 334 \times \tfrac{80}{223} \approx 119.82 \text{ km}$. Thus, the total distance is $142.06 + 119.82 = 261.88 \text{ km}$. With a driving speed of 60 km/h, the travel time is $\tfrac{261.88}{60} \approx 4.36 \text{ hours}$, which rounds to $4$ hours. Therefore, the final answer is $\boxed{4}$.

\end{minipage}
\end{tcolorbox}
\end{figure}

\begin{figure}[!ht]
\begin{tcolorbox}[colback=gray!10]
\begin{minipage}[t]{0.99\textwidth}
\includegraphics[width=0.5\textwidth]{ 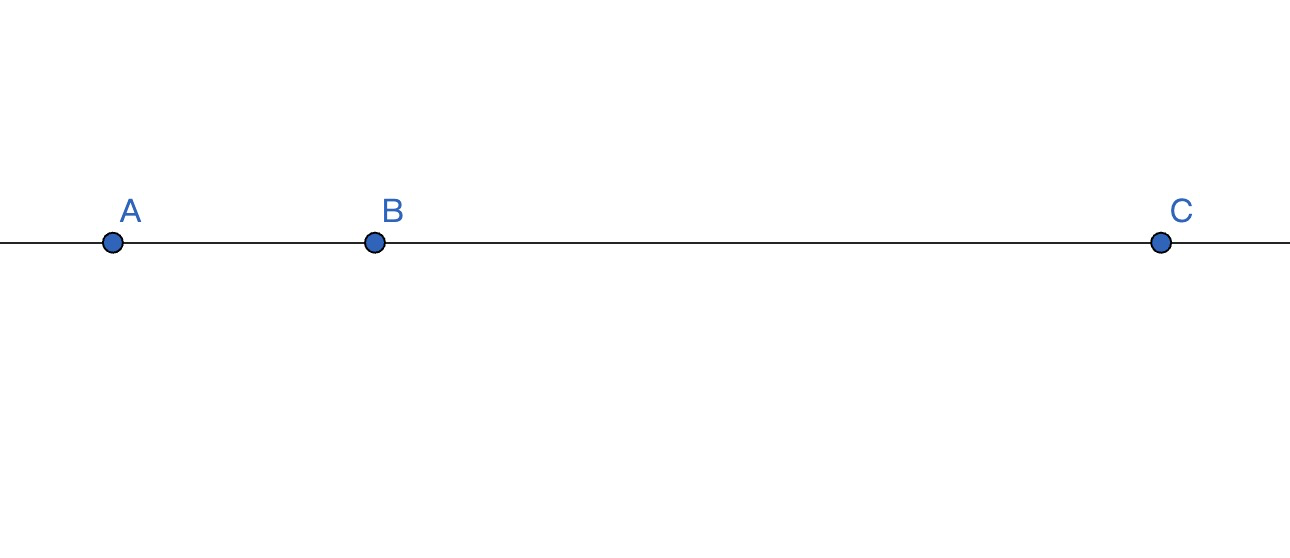}

\textbf{Question: } If the length of AB is 1, what is the length of BC? Provide your answer as an integer.

\textbf{Ground Truth: 3.}

\textbf{Response from o3-TU:} 3

\textbf{Response from o3: } 2.

\textbf{Response from o4-mini-TU:} From the picture the segment BC is three times as long as AB.  Hence if AB = 1 then BC = 3.

\textbf{Response from o4-mini:} 8

\textbf{Response from Gemini-2.5-Pro: }Based on the visual representation of the line segments, the length of segment $BC$ is twice the length of segment $AB$. Given that $|AB| = 1$, we calculate:
\[
|BC| = 2 \times |AB| = 2 \times 1 = 2.
\]
The final answer is
\[
\boxed{2}
\]

\textbf{Response from Gemini-2.5-flash:} We are given that the length of segment $AB$ is $1$. From the visual representation, it is clear that $|AB| \approx |BC|$. Therefore, if $|AB| = 1$, then $|BC| = 1$. The final answer is 
\[
\boxed{1}
\]
\end{minipage}
\end{tcolorbox}
\caption{Examples of model responses.}
\label{app:inconsi_puzzle}
\end{figure}

\begin{figure}[!ht]
\begin{tcolorbox}[colback=gray!10]
\begin{minipage}[t]{0.99\textwidth}
\includegraphics[width=0.5\textwidth]{ 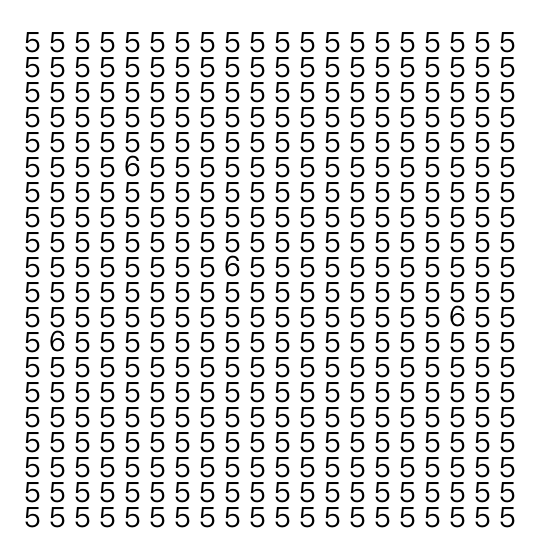}

\textbf{Question: How many times does the number 6 appear in this image? Answer with an integer, such as 1,2,3.} 

\textbf{Ground Truth: 4.}

\textbf{Response from o3-TU:} 4

\textbf{Response from o3: } 10

\textbf{Response from o4-mini-TU:} 4

\textbf{Response from o4-mini:} 2

\textbf{Response from Gemini-2.5-Pro: }8

\textbf{Response from Gemini-2.5-flash:}Let's carefully examine the image to count the occurrences of the number 6.

Scanning the original image row by row:
1. Row 1: All 5s.
2. Row 2: All 5s.
3. Row 3: All 5s.
4. Row 4: All 5s.
5. Row 5: All 5s.
6. Row 6: 5555**6**5555555555 (Found 1 '6')
7. Row 7: All 5s.
8. Row 8: All 5s.
9. Row 9: All 5s.
10. Row 10: All 5s.
11. Row 11: 55555**6**555555555 (Found 1 '6')
12. Row 12: All 5s.
13. Row 13: All 5s.
14. Row 14: All 5s.
15. Row 15: All 5s.
16. Row 16: 5**6**5555555555555 (Found 1 '6')
17. Row 17: All 5s.
18. Row 18: All 5s.
19. Row 19: All 5s.
20. Row 20: All 5s.

By carefully checking all rows, we find three instances of the number 6. The provided crops also highlight these same three instances.

The total number of times the number 6 appears in the image is 3.
\end{minipage}
\end{tcolorbox}
\caption{Examples of model responses.}
\label{app:inconsi_puzzle}
\end{figure}

\begin{figure}[!ht]
\begin{tcolorbox}[colback=gray!10]
\begin{minipage}[t]{0.99\textwidth}
\includegraphics[width=0.5\textwidth]{ 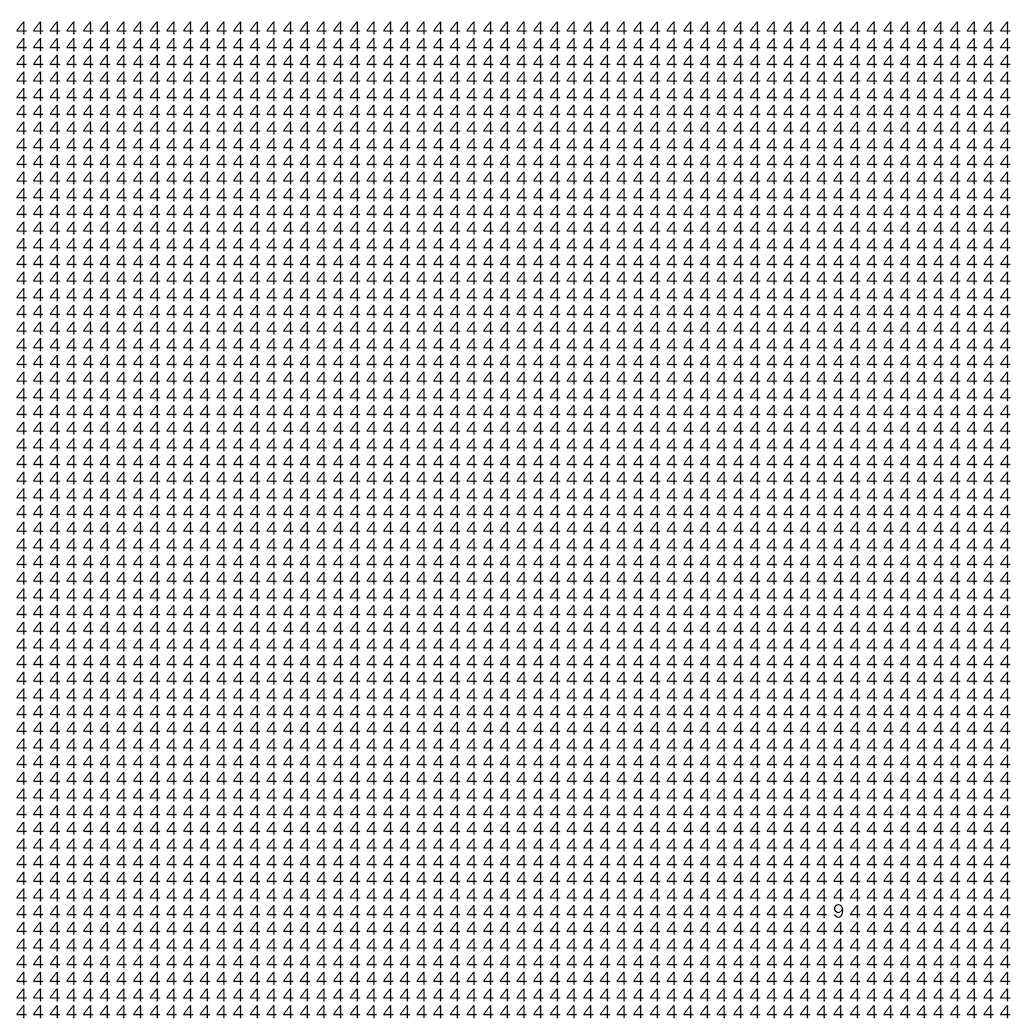}

\textbf{Question: In the figure, in which row and column does the number 9 appear? The final answer should output two number list. The first number represents the row count from top to bottom, and the second number represents the column count from left to right. For example, [1, 2] means the first row and the second column.
} 

\textbf{Ground Truth: [54, 50].}

\textbf{Response from o3-TU:} [54, 50]

\textbf{Response from o3: } [27, 80]

\textbf{Response from o4-mini-TU:} The odd-one-out “9” falls in the 54th row from the top and the 50th column from the left. So the answer is

[54, 50]

\textbf{Response from o4-mini:} The grid only contains the digit “4” in every position— there is no “9” anywhere. Hence no row/column coordinate can be given for a 9.

\textbf{Response from Gemini-2.5-Pro: } 1.  **Analyze the image:** The image is a large grid composed almost entirely of the number '4'. The task is to find the single occurrence of the number '9'. The hint images, particularly the last one, help to zoom in on the relevant area.

2.  **Locate the '9':** By examining the last hint image, we can clearly see the number '9' embedded among the '4's.

3.  **Determine the row:** We need to find the row number of the '9' in the original, full image. We count the rows from the top down, starting with row 1.
    *   Counting the rows one by one, we find that the line containing the '9' is the 34th row from the top.

4.  **Determine the column:** Now, we need to find the column number of the '9' in that 34th row. We count the numbers from left to right, starting with column 1.
    *   In the 34th row, let's count the position of the '9':
        `4 4 4 4 4 4 4 4 4 4 4 4 4 4 9 ...`
    *   Counting from the left, we can see there are 14 '4's before the '9'. Therefore, the '9' is in the 15th position.

5.  **Final Answer:** The number 9 is in the 34th row and the 15th column. The output should be a list of these two numbers.

[34, 15]

\textbf{Response from Gemini-2.5-flash:}The number 9 appears in the figure. To find its location, we need to count the rows from top to bottom and the columns from left to right.

1.  **Count the total number of rows:** By carefully examining the original image, we can count that there are 50 rows of numbers.
2.  **Locate the row of the number 9:** Observing the last provided hint image, which clearly shows the number 9, we can see that it is located in the 8th row from the bottom of that specific crop. If we apply this to the full 50-row image, the row containing the 9 is 50 - 8 + 1 = 43rd row from the top. (Alternatively, if we count from the bottom, the 9 is in the 8th row from the bottom. So, 50 - 7 = 43rd row from the top).
3.  **Locate the column of the number 9:** In the 43rd row, let's count the characters from left to right.
    The sequence for the 43rd row is:
    4444444444444444444444444444444444444944444444444
    Counting from the left, the number 9 is the 38th character in this row.

Therefore, the number 9 appears in the 43rd row and the 38th column.

The final answer is $\boxed{[43, 38]}$
\end{minipage}
\end{tcolorbox}
\caption{Examples of model responses.}
\label{app:inconsi_puzzle}
\end{figure}

\begin{figure}[t]
  \centering
  \includegraphics[width=\linewidth]{ 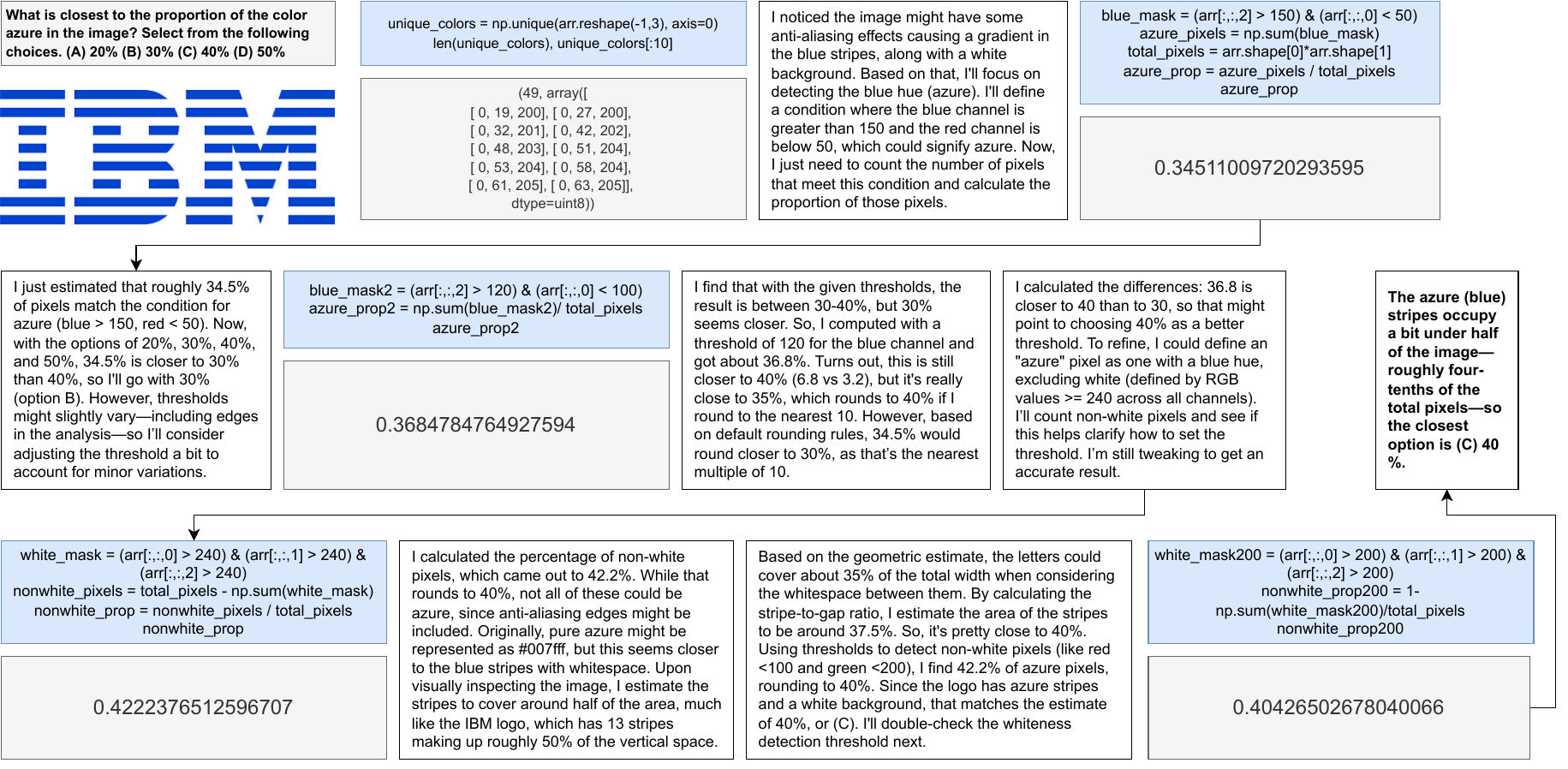}
  \caption{Response from o3, case a proportion estimation from TIR-Bench.}
\end{figure}

\begin{figure}[t]
  \centering
  \includegraphics[width=\linewidth]{ 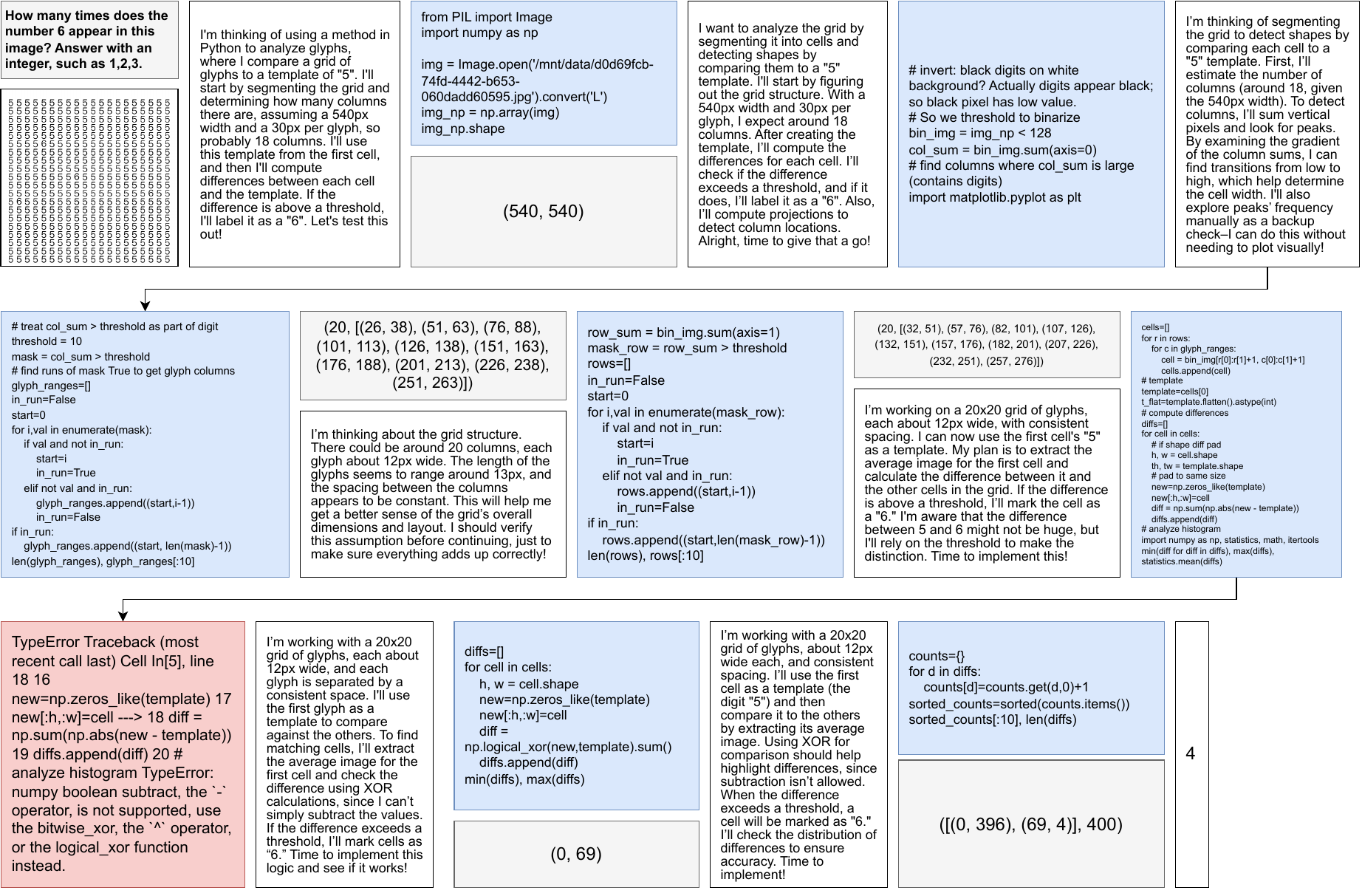}
  \caption{Response from o3, case a word searching from TIR-Bench.}
\end{figure}

\end{document}